\newif\ifprl
\ifprl
\documentclass[times,twocolumn,final,authoryear]{elsarticle}
\else
\documentclass[10pt]{article}
\newcommand\citep[1]{\cite{#1}}
\fi

\newcommand\dome{$DOME$\@\xspace}
\newcommand\pdome{$PDOME$\@\xspace}
\newcommand\mdome{$MDOME$\@\xspace}

\ifprl
\usepackage{prletters}
\else
\usepackage{layout}
\usepackage[margin=60pt]{geometry}
\fi
\usepackage{framed,multirow}
\usepackage{amssymb}
\usepackage{latexsym}
\usepackage{amsmath}
\usepackage[capitalise]{cleveref}
\usepackage{booktabs}       

\usepackage{url}
\usepackage{xcolor}
\definecolor{newcolor}{rgb}{.8,.349,.1}

\usepackage{graphicx}
\usepackage[caption=false,font=footnotesize]{subfig}
\usepackage{float}

\newcommand{\norm}[1]{\left\lVert #1 \right\rVert}

\usepackage{xspace}
\providecommand{\eg}[0]{e.g.,\@\xspace}

\usepackage{setspace}
\singlespace

\ifprl
\journal{Pattern Recognition Letters}
\fi

\begin{document}

\ifprl
\thispagestyle{empty}
                                                             
\begin{table*}[!th]

\begin{minipage}{.9\textwidth}
\baselineskip12pt
\ifpreprint
  \vspace*{1pc}
\else
  \vspace*{-6pc}
\fi

\noindent {\LARGE\itshape Pattern Recognition Letters}
\vskip6pt

\noindent {\Large\bfseries Authorship Confirmation}

\vskip1pc

{\bf Please save a copy of this file, complete and upload as the 
``Confirmation of Authorship'' file.}

\vskip1pc

As corresponding author 
I, \underline{\hphantom{\hspace*{7cm}}}, 
hereby confirm on behalf of all authors that:

\vskip1pc

\begin{enumerate}
\itemsep=3pt
\item This manuscript, or a large part of it, \underline {has not been
published,  was not, and is not being submitted to} any other journal. 

\item If \underline {presented} at or \underline {submitted} to or
\underline  {published }at a conference(s), the conference(s) is (are)
identified and  substantial \underline {justification for
re-publication} is presented  below. A \underline {copy of
conference paper(s) }is(are) uploaded with the  manuscript.

\item If the manuscript appears as a preprint anywhere on the web, e.g.
arXiv,  etc., it is identified below. The \underline {preprint should
include a  statement that the paper is under consideration at Pattern
Recognition  Letters}.

\item All text and graphics, except for those marked with sources, are
\underline  {original works} of the authors, and all necessary
permissions for  publication were secured prior to submission of the
manuscript.

\item All authors each made a significant contribution to the research
reported  and have \underline {read} and \underline {approved} the
submitted  manuscript. 
\end{enumerate}

Signature\underline{\hphantom{\hspace*{7cm}}} Date\underline{\hphantom{\hspace*{4cm}}} 
\vskip1pc

\rule{\textwidth}{2pt}
\vskip1pc

{\bf List any pre-prints:}
\vskip5pc

\rule{\textwidth}{2pt}
\vskip1pc

{\bf Relevant Conference publication(s) (submitted, accepted, or
published):}
\vskip5pc

{\bf Justification for re-publication:}

\end{minipage}
\end{table*}

\clearpage
\thispagestyle{empty}
\ifpreprint
  \vspace*{-1pc}
\fi

\begin{table*}[!th]
\ifpreprint\else\vspace*{-5pc}\fi

\section*{Graphical Abstract (Optional)}
To create your abstract, please type over the instructions in the
template box below.  Fonts or abstract dimensions should not be changed
or altered. 

\vskip1pc
\fbox{
\begin{tabular}{p{.4\textwidth}p{.5\textwidth}}
\bf Under the DOME of a compactness-inducing activation function  \\
Mohamed E. Hussein and Wael Abd-Almageed
\includegraphics[width=.3\textwidth]{top-elslogo-fm1.pdf}
& 
This is the dummy text for graphical abstract.
This is the dummy text for graphical abstract.
This is the dummy text for graphical abstract.
This is the dummy text for graphical abstract.
This is the dummy text for graphical abstract.
This is the dummy text for graphical abstract.
This is the dummy text for graphical abstract.
This is the dummy text for graphical abstract.
This is the dummy text for graphical abstract.
This is the dummy text for graphical abstract.
This is the dummy text for graphical abstract.
This is the dummy text for graphical abstract.
This is the dummy text for graphical abstract.
This is the dummy text for graphical abstract.
This is the dummy text for graphical abstract.
This is the dummy text for graphical abstract.
This is the dummy text for graphical abstract.
This is the dummy text for graphical abstract.
This is the dummy text for graphical abstract.
This is the dummy text for graphical abstract.
This is the dummy text for graphical abstract.
This is the dummy text for graphical abstract.
This is the dummy text for graphical abstract.
This is the dummy text for graphical abstract.
\end{tabular}
}

\end{table*}

\clearpage
\thispagestyle{empty}

\ifpreprint
  \vspace*{-1pc}
\else
\fi

\begin{table*}[!t]
\ifpreprint\else\vspace*{-15pc}\fi

\section*{Research Highlights (Required)}

To create your highlights, please type the highlights against each
\verb+\item+ command. 

\vskip1pc

\fboxsep=6pt
\fbox{
\begin{minipage}{.95\textwidth}
It should be short collection of bullet points that convey the core
findings of the article. It should  include 3 to 5 bullet points
(maximum 85 characters, including spaces, per bullet point.)  
\vskip1pc
\begin{itemize}

 \item 

 \item 

 \item

 \item

 \item

\end{itemize}
\vskip1pc
\end{minipage}
}

\end{table*}

\clearpage

\ifpreprint
  \setcounter{page}{1}
\else
  \setcounter{page}{1}
\fi

\begin{frontmatter}

\title{Under the DOME of a compactness-inducing activation function}

\author[1]{Mohamed E. Hussein\corref{cor1}%
\fnref{fn1}}
\ead{mehussein@isi.edu}
\author[1]{Wael Abd-Almageed\fnref{fn2}}
\ead{wamageed@isi.edu}

\cortext[cor1]{Corresponding author}
\fntext[fn1]{Also, Associate Professor on leave from Alexandria University, Alexandria, Egypt.}
\fntext[fn2]{Also, Research Associate Professor at USC Viterbi School of Engineering.}
\address[1]{Information Sciences Institute, University of Southern California, Marina del Rey, CA 90292, USA}

\received{01 OCT 2021}
\finalform{DD MON 2021}
\accepted{DD MON 2021}
\availableonline{DD Mon 2021}
\communicated{}

\begin{abstract}
In this paper, we introduce a novel non-linear activation function that spontaneously induces class-compactness and regularization in the embedding space of neural networks. The function is dubbed DOME for \emph{Difference Of Mirrored Exponential terms}. The basic form of the function can replace the sigmoid or the hyperbolic tangent functions as an output activation function for binary classification problems. The function can also be extended to the case of multi-class classification, and used as an alternative to the standard softmax function. It can also be further generalized to take more flexible shapes suitable for intermediate layers of a network.
We evaluate the DOME activation on multiple classification tasks, showing that DOME surpasses the baseline softmax' performance. Moreover, we show that the function leads to enhanced robustness against adversarial attacks. The benefits of DOME are further boosted when its generalized variant is applied to the intermediate layers of classification models.
\end{abstract}

\begin{keyword}
non-linear activation function, softmax, face recognition, adversarial robustness

\end{keyword}

\end{frontmatter}

\else 
\title{Introducing the DOME Activation Functions}

\author{
Mohamed E. Hussein \\ Information Sciences Institute\\University of Southern California\\Marina del Rey, CA 90292, USA
\and
Wael AbdAlmageed \\ Information Sciences Institute\\University of Southern California\\Marina del Rey, CA 90292, USA
}

\date{}

\maketitle

\begin{abstract}

In this paper, we introduce a novel non-linear activation function that spontaneously induces class-compactness and regularization in the embedding space of neural networks. The function is dubbed DOME for \emph{Difference Of Mirrored Exponential terms}. The basic form of the function can replace the sigmoid or the hyperbolic tangent functions as an output activation function for binary classification problems. The function can also be extended to the case of multi-class classification, and used as an alternative to the standard softmax function. It can also be further generalized to take more flexible shapes suitable for intermediate layers of a network.
We empirically demonstrate the properties of the function. We also show that models using the function exhibit extra robustness against adversarial attacks.

\end{abstract}

\fi 


\section{Introduction}
\label{sec:intro}
Over the past decade, numerous breakthroughs in artificial intelligence (AI) have been made possible by advances in deep neural networks \citep{krizhevskyImageNetClassificationDeep2017,qianVeryDeepConvolutional2016,vinyalsShowTellLessons2017,zhangDeepNeuralNetworks2015,Brown2020LanguageMA,Karras2019stylegan2}. Such advances spanned many aspects of training and evaluating deep neural networks, such as model architectures \citep{ioffeBatchNormalizationAccelerating2015,srivastavaDropoutSimpleWay2014,He_2016_CVPR,vaswaniAttentionAllYou2017a,sabourDynamicRoutingCapsules2017}, weight initialization \citep{glorotUnderstandingDifficultyTraining2010a,heDelvingDeepRectifiers2015a,klambauerSelfnormalizingNeuralNetworks2017}, activation functions \citep{nairRectifiedLinearUnits2010a,clevert2016elu}, optimization algorithms \citep{kingmaAdamMethodStochastic2015,luoAdaptiveGradientMethods2018}, data preparation \citep{luoAdaptiveGradientMethods2018,hendrycks*AugMixSimpleData2019b}, and training paradigms \citep{hintonDistillingKnowledgeNeural2015,goodfellowGenerativeAdversarialNets2014,zhuUnpairedImageToImageTranslation2017,devlinBERTPretrainingDeep2019a,heMomentumContrastUnsupervised2020a}. Parallel to these advances, the threat posed by adversarial examples to machine learning models was also discovered \citep{biggioEvasionAttacksMachine2013b,brunaIntriguingPropertiesNeural2014}, and has been attracting much attention in the research community \citep{xuAdversarialAttacksDefenses2020a,chakrabortySurveyAdversarialAttacks2021}.

In this paper, we introduce a novel non-linear activation function dubbed \dome, for Difference Of Mirrored Exponential terms. \dome is very distinctive in its shape and effect from any other activation function in the literature. \dome is a continuously differentiable, aperiodic, and non-monotonic activation function. In its simplest form, \dome is similar to the sigmoid or hyperbolic tangent functions, which are common choices for output activation functions in binary classification problems. However, different from both of these functions, \dome naturally induces compactness of each class and separation between different classes in the embedding space. \dome can be parameterized to provide a more general form that can be used in any layer throughout a neural network. It can also be generalized to substitute the softmax function as an output activation for multi-class classification problems. However, unlike softmax, \dome naturally induces intra-class compactness and inter-class separation in the embedding space without the addition of special losses. \dome also enjoys a self-regularization property.

We empirically demonstrate the effectiveness of the \dome function in different classification tasks. \dome's advantage is most evident in the extra robustness it adds against adversarial examples. 

In the rest of this paper, we first give an overview of relevant literature (\cref{sec:rel_wrk}). Next, we lay the backdrop with the simple case of binary classification (\cref{sec:act_func}). Afterwards, we introduce the \dome function (\cref{sec:dome}), its generalized version (\cref{sec:pdome}), and its multi-class extension (\cref{sec:mdome}).
Finally, we present our empirical evaluation and analysis (\cref{sec:exps}) before concluding the paper (\cref{sec:conc}).

\section{Related work}
\label{sec:rel_wrk}
In this section, we discuss the most relevant prior research to ours.

\subsection{Non-linear activation functions}
\label{sec:rw_non_linearities}
Saturating activation functions, \eg the logistic sigmoid and the hyperbolic tangent, used to be the dominant non-linearities applied in neural networks. This has changed since the rectified linear units (ReLUs) was found to be more effective in training deep neural networks \citep{nairRectifiedLinearUnits2010a}. Since then, many more non-linear activation functions have been introduced in the literature. The effectiveness of ReLUs was partly believed to be due to the sparseness it induces on network activations. However, this belief was later contradicted by the success of non-sparsity-inducing activation functions. For example, the leaky ReLU function was introduced to facilitate the gradient propagation with ReLUs by introducing a non-zero slope in the negative part of the function's domain \citep{maasRectifierNonlinearitiesImprove2013a}, which essentially deprived ReLU from it sparseness effect. Leaky ReLUs were further generalized to parametric ReLUs \citep{heDelvingDeepRectifiers2015a}, in which the slope of the negative side was turned into a learnable parameter, and randomized ReLUs \citep{xuEmpiricalEvaluationRectified2015a}, in which that slope was made randomized. All of these ReLU variants overcame the lack of gradient in the negative part of the domain, which hampered the speed of convergence in the original ReLU.

The exponential linear units (ELUs) \citep{clevert2016elu} was then introduced to maintain the non-zero gradient in the negative part of the domain while making the function saturate in that part. Different from all the preceding functions, the continuously differentiable exponential linear units (CELUs) \citep{barronContinuouslyDifferentiableExponential2017}, as its name indicates, is continuously differentiable while still maintaining all the desired properties of ReLUs and ELUs. The Gaussian error linear units (GELUs) \citep{hendrycksGaussianErrorLinear2018} and the Swish function \citep{ramachandranSearchingActivationFunctions2017} were then introduced to add the extra generality of not being monotonic in the negative part of their domains. Another interesting tweak of ELUs was the scaled ELUs (SELUs) \citep{klambauerSelfnormalizingNeuralNetworks2017}, which was shown to introduced a self-normalization feature that could eliminate the need to use batch normalization layers \citep{ioffeBatchNormalizationAccelerating2015}. The sinosoid activation function was also shown to be amenable to train deep networks despite the multiple local modes \citep{parascandoloTamingWavesSine2016}. Our proposed activation function is most similar to the sinosoid, without being periodic. \dome is also non-monotonic and continuously differentiable.

The aformentioned activation functions are designed with at least one of these goals in mind: breaking the linearity of the model, inducing sparsity, facilitating gradient computation and propagation, being biologically inspired, or inducing compactness. It is worth noting that there is another class of activation functions designed with the goal of flexibility to represent a very broad class of functions, such as the cubic spline activation \citep{scardapaneLearningActivationFunctions2019} and the Maxout function \citep{goodfellowMaxoutNetworks2013}.

\subsection{Priors on embedding spaces}
\label{sec:rw_priors}
The arrangement of points in the embedding space of a neural network has a significant effect on its performance on different tasks. There is a sizeable body of research that enforces different priors on that space to encourage a certain behavior.

\cite{liuLearningMinimumHyperspherical2018} applied hyperspherical energy regularization to minimize redundancy between features of the same layer. In the output layer, this approach resulted in encouraging class centers to be as spread as possible. \cite{sablayrollesSpreadingVectorsSimilarity2018} applied a similarly formulated regularization that was inspired by maximizing the differential entropy. The notion of enforcing class-compactness in the embedding space has attracted much attention in the face recognition community. Typically, such proposed methods were posed either as an added loss function, such as the center loss \citep{wenDiscriminativeFeatureLearning2016,wenComprehensiveStudyCenter2019}; as an added constraint, such as the ring loss \citep{zhengRingLossConvex2018} and crystal loss \citep{ranjanCrystalLossQuality2019,ranjanL2constrainedSoftmaxLoss2017}; or a reformulation of the cross entropy loss, such as in Gaussian-mixture loss \citep{wanRethinkingFeatureDistribution2018a}, angular softmax \citep{liuSphereFaceDeepHypersphere2017}, NormFace \citep{wangNormFaceL2Hypersphere2017}, CosFace \citep{wangCosFaceLargeMargin2018}, ArcFace \citep{dengArcFaceAdditiveAngular2019}. It is worth noting that the aforementioned reformulations of the cross entropy loss, despite being posed as such, can be seen as reformulations of the penultimate layer's function without introducing changes to the softmax function or the cross entropy loss.


\subsection{Defenses against adversarial attacks}
\label{sec:rw_defenses}
Since the discovery of adversarial attacks against machine learning models, many attempts have been introduced to defend against them. The most successful defense so far is adversarial training and its variants \citep{madryDeepLearningModels2018,zhangTheoreticallyPrincipledTradeoff2019}. Similar to the \dome function as a defense, there were other attempts to induce compactness in the embedding space as a defense mechanism \citep{pangMaxMahalanobisLinearDiscriminant2018b,pangRethinkingSoftmaxCrossEntropy2019}. However, as many other defenses that did not involve adversarial training \citep{bafnaThwartingAdversarialExamples2018,pangImprovingAdversarialRobustness2019a,xiaoEnhancingAdversarialDefense2019}, such approaches were soon defeated via adaptive attacks \citep{tramerAdaptiveAttacksAdversarial2020} or automatically adjustable attack recipes \citep{croceReliableEvaluationAdversarial2020b}. We are not proposing the \dome function as a standalone defense. The \dome function constitutes a defense when combined with adversarial training.

\section{Binary classification output activation}

\label{sec:act_func}

\begin{figure*}[ht]
	\centering
	\subfloat[Classification model]
	{
		\includegraphics[width=0.27\linewidth]{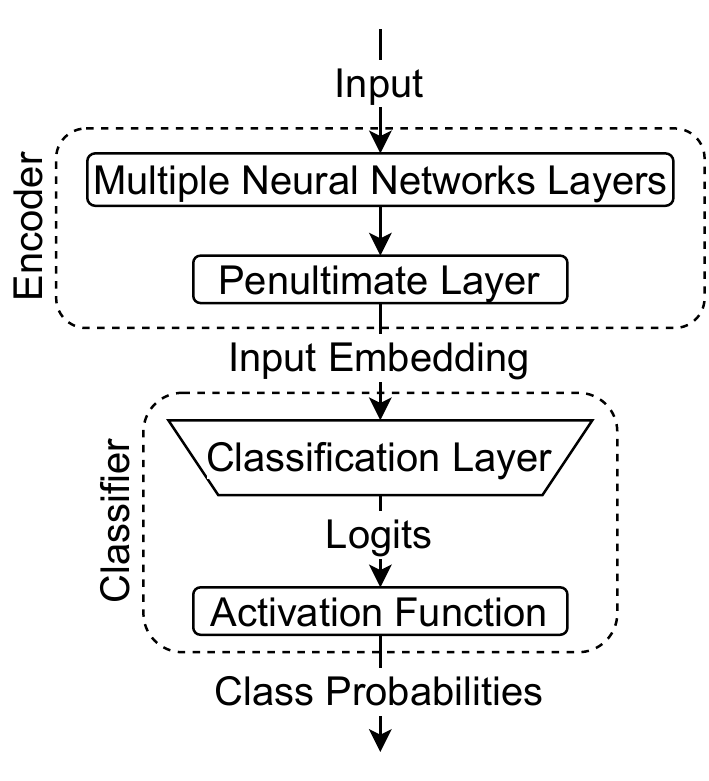}
		\label{fig:nn_model}
	}
	\subfloat[$\tanh(x)=\frac{e^x-e^{-x}}{e^x+e^{-x}}$]
	{
		\includegraphics[width=0.35\linewidth]{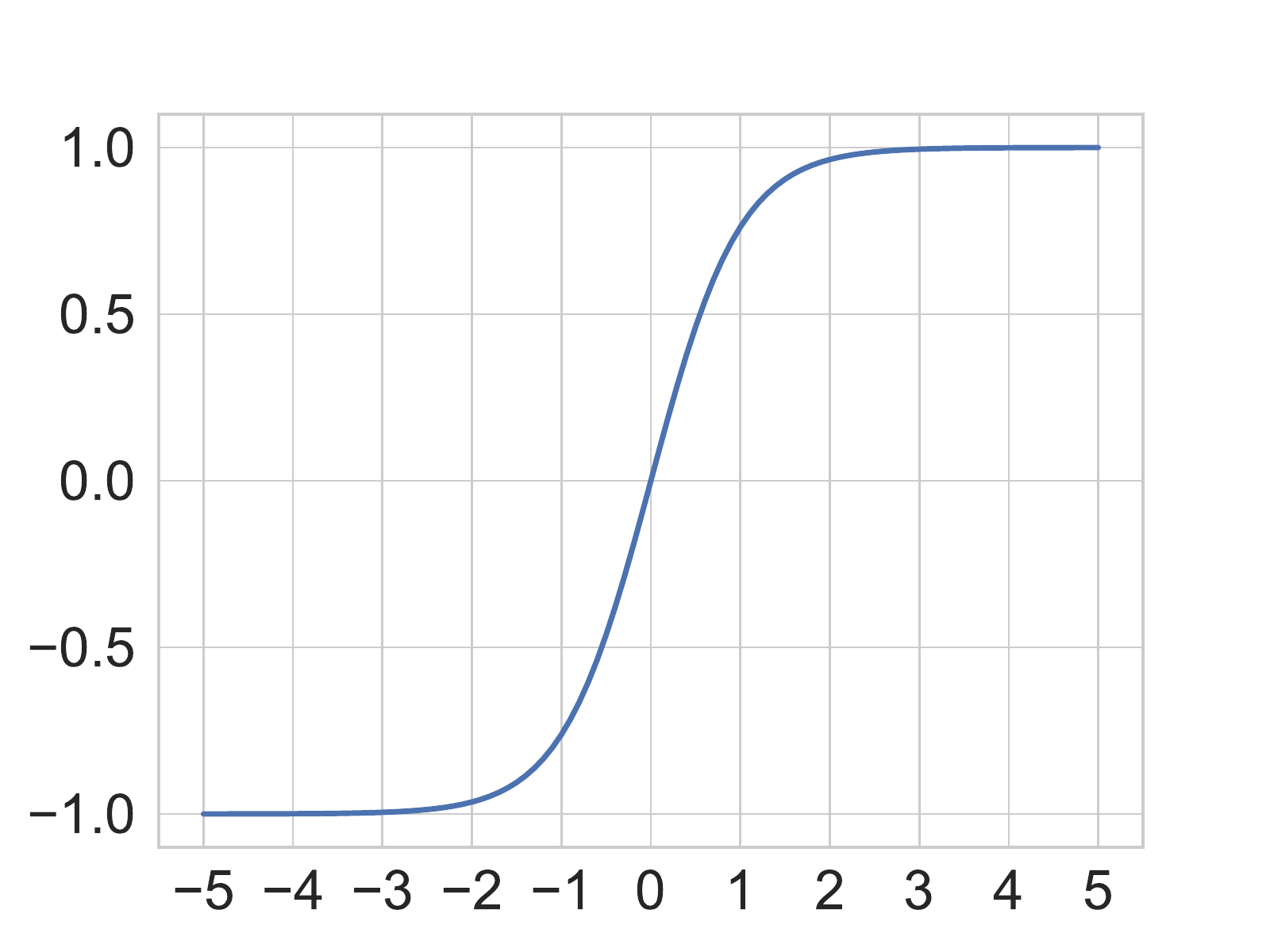}
		\label{fig:tanh_func}
	}
	\subfloat[A linear output layer for a 2D embedding space]
	{
		\includegraphics[width=0.25\linewidth]{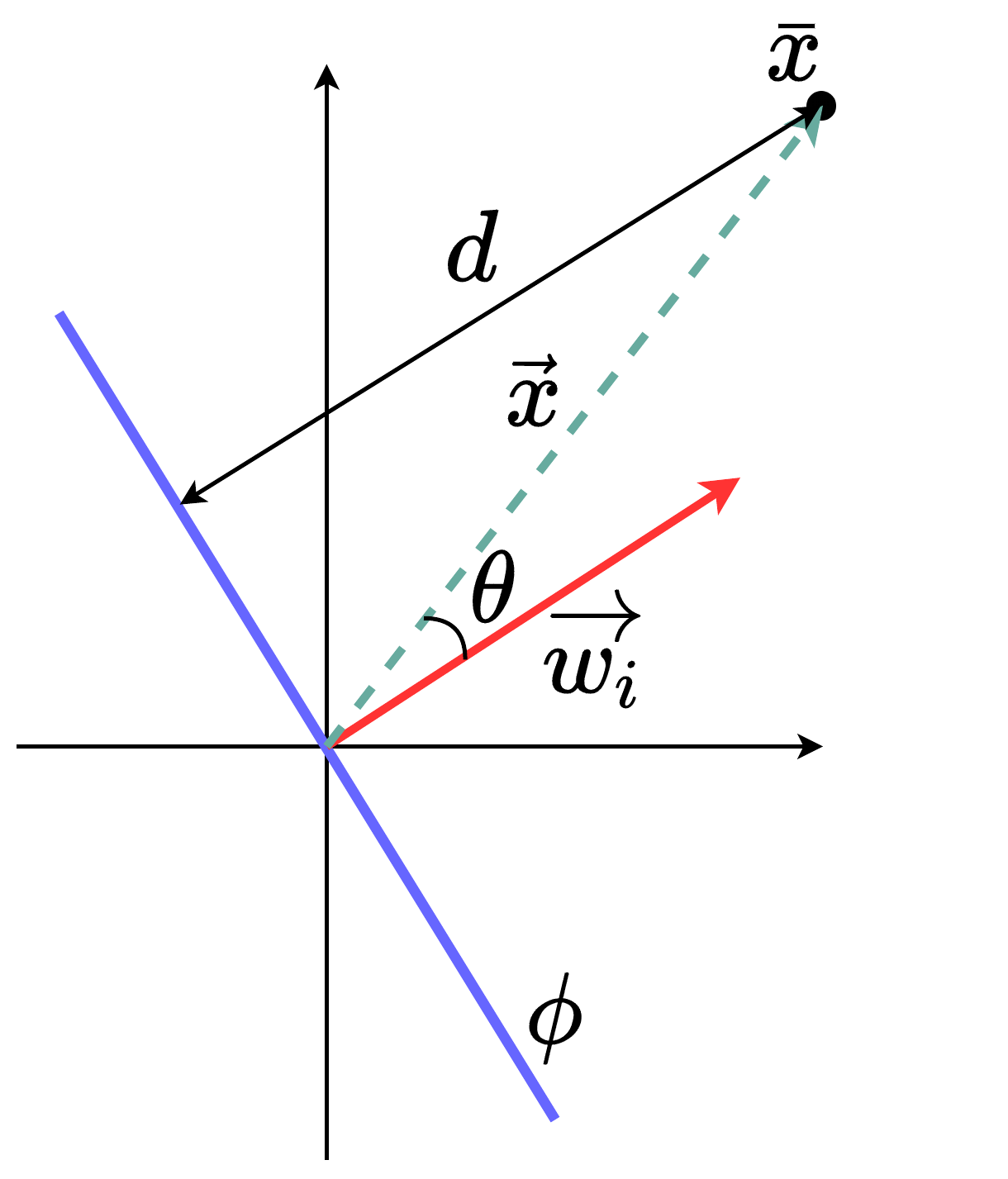}
        \label{fig:2d_logit}
	}
	\caption{Structure of a generic classification neural network model, the tanh activation functions, and an illustration of a 2D embedding space with a classification hyperplane.}
	\label{fig:nn_model_tanh_logits}
\end{figure*}

\Cref{fig:nn_model} shows the structure of a generic neural network model. The network consists of two main parts, the \textit{encoder} and the \textit{classifier}. The last layer of the encoder is the \textit{penultimate layer}, whose output constitutes the input's \textit{embedding}. The classifier relies on the embedding to produce the classification outputs. The classification stage can be split into two sub-stages, one that produces real valued \textit{logits}, and the other maps the logits into a probability distribution over the possible output classes through a non-linear \textit{activation function}.

\begin{figure*}[ht]
	\centering
	\subfloat[$\mbox{sigmoid}(x)=\frac{1}{1+e^{-x}}$]
	{
		\includegraphics[width=0.3\linewidth]{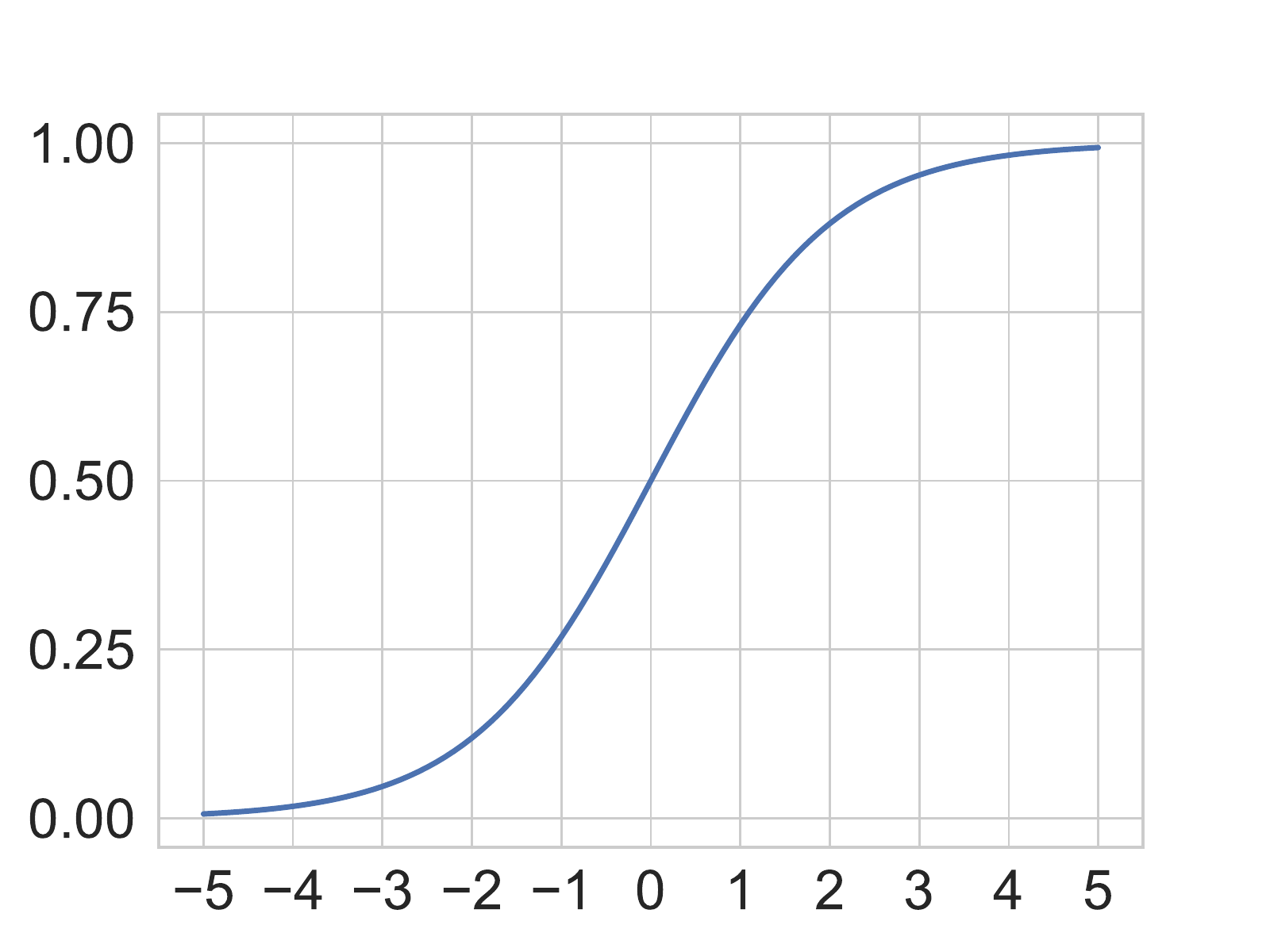}
		\label{fig:sigmoid_func}
	} 
	\subfloat[$\mbox{softmax}(x)=\frac{e^x}{e^x+e^{-x}}$]
	{
		\includegraphics[width=0.3\linewidth]{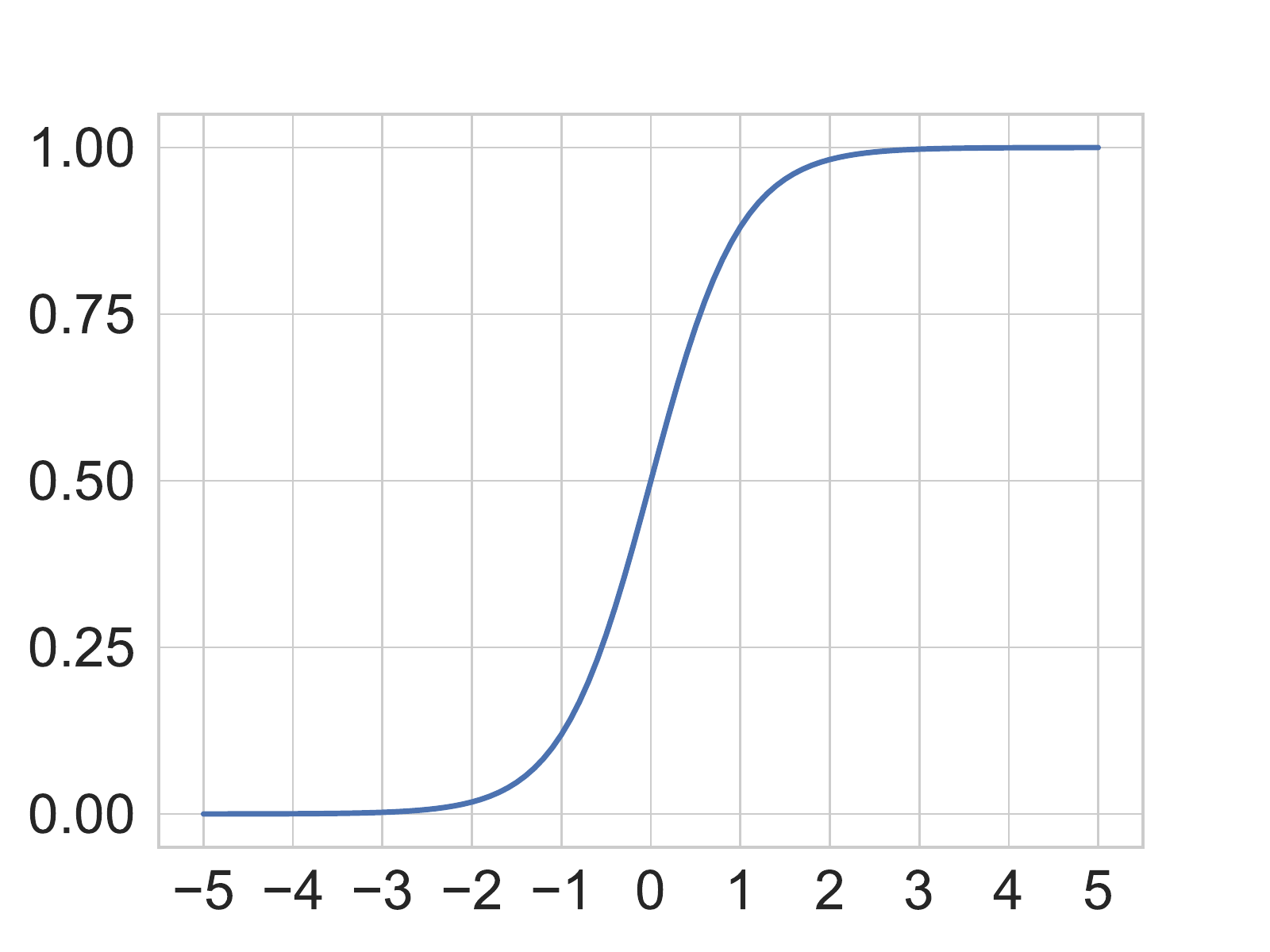}
		\label{fig:softmax_func}
	}
	\subfloat[The \dome function]
	{
		\includegraphics[width=0.3\linewidth]{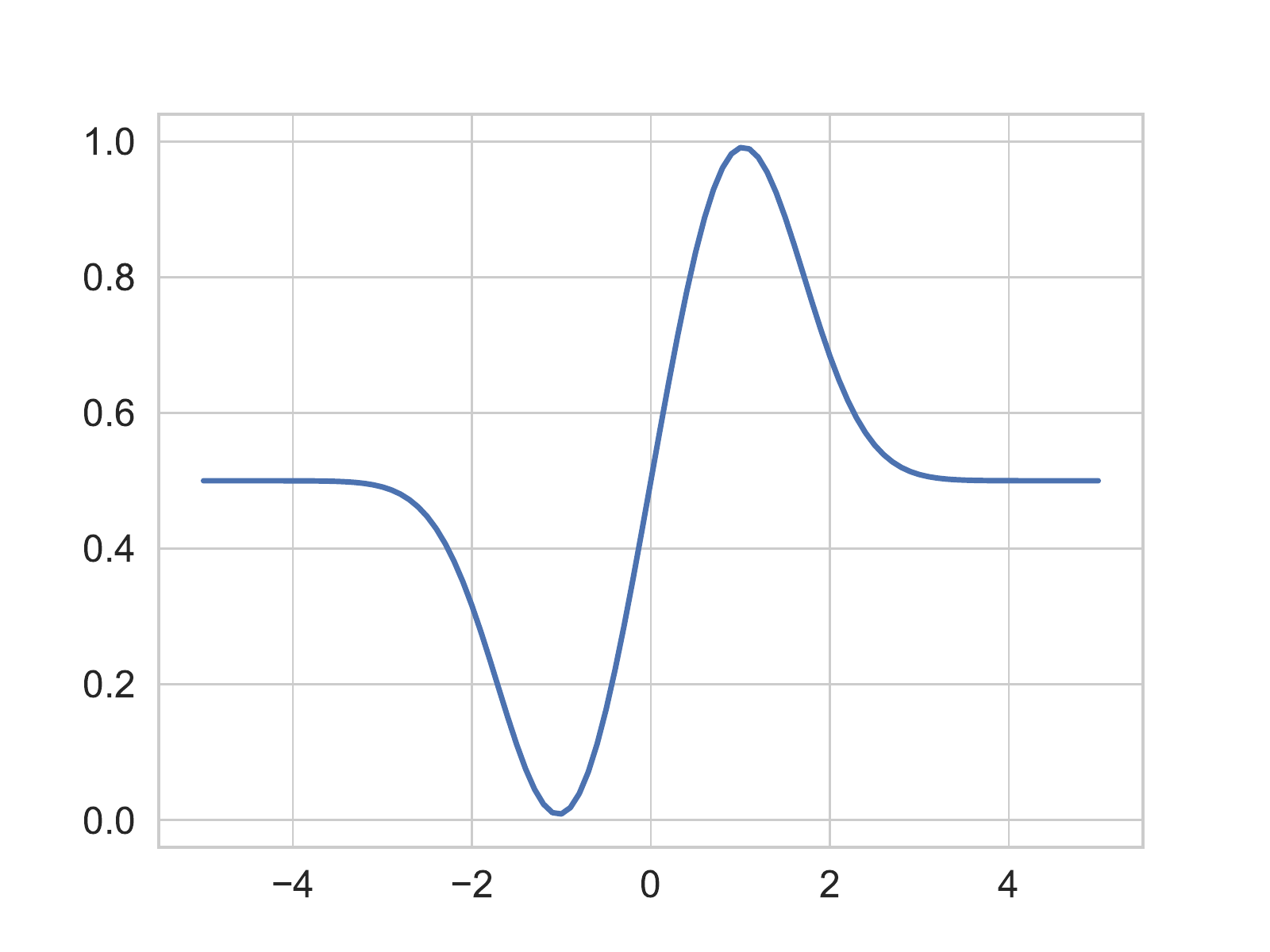}
		\label{fig:dome_func}
	}
	\caption{Plots for three activation function, the sigmoid, the softmax, and the proposed \dome.}
	\label{fig:functions}
\end{figure*}

Let's first have a look at the effect of output activation functions on the embedding space in the simple case of binary classification. Binary classification neural network models use one or two output units with a non-linear activation function. \Cref{fig:tanh_func,fig:sigmoid_func,fig:softmax_func} display three classic activation functions commonly used in this case. If the model has a single output unit, the activation function is typically either the tanh function (\cref{fig:tanh_func}) or the sigmoid function (\cref{fig:sigmoid_func}), whose values are always in the intervals $(-1, 1)$ or $(0, 1)$, respectively, for any finite input. If the model has two output units, the softmax function is the natural choice. The softmax function, in this case, has two outputs, each of which is in $(0, 1)$ and their sum is $1$. \Cref{fig:softmax_func} shows one of the two outputs in the special case of having the two inputs equal in value and opposite in sign. This special case of softmax is used here for illustration only. However, the phenomenon we are about to discuss applies to the general case as well. Note that if one of the inputs to the softmax function is zero, the function reduces to a sigmoid function of the other input.

\begin{figure*}[ht]
	\centering
	\subfloat[Sigmoid]
	{
		\includegraphics[width=0.3\linewidth]{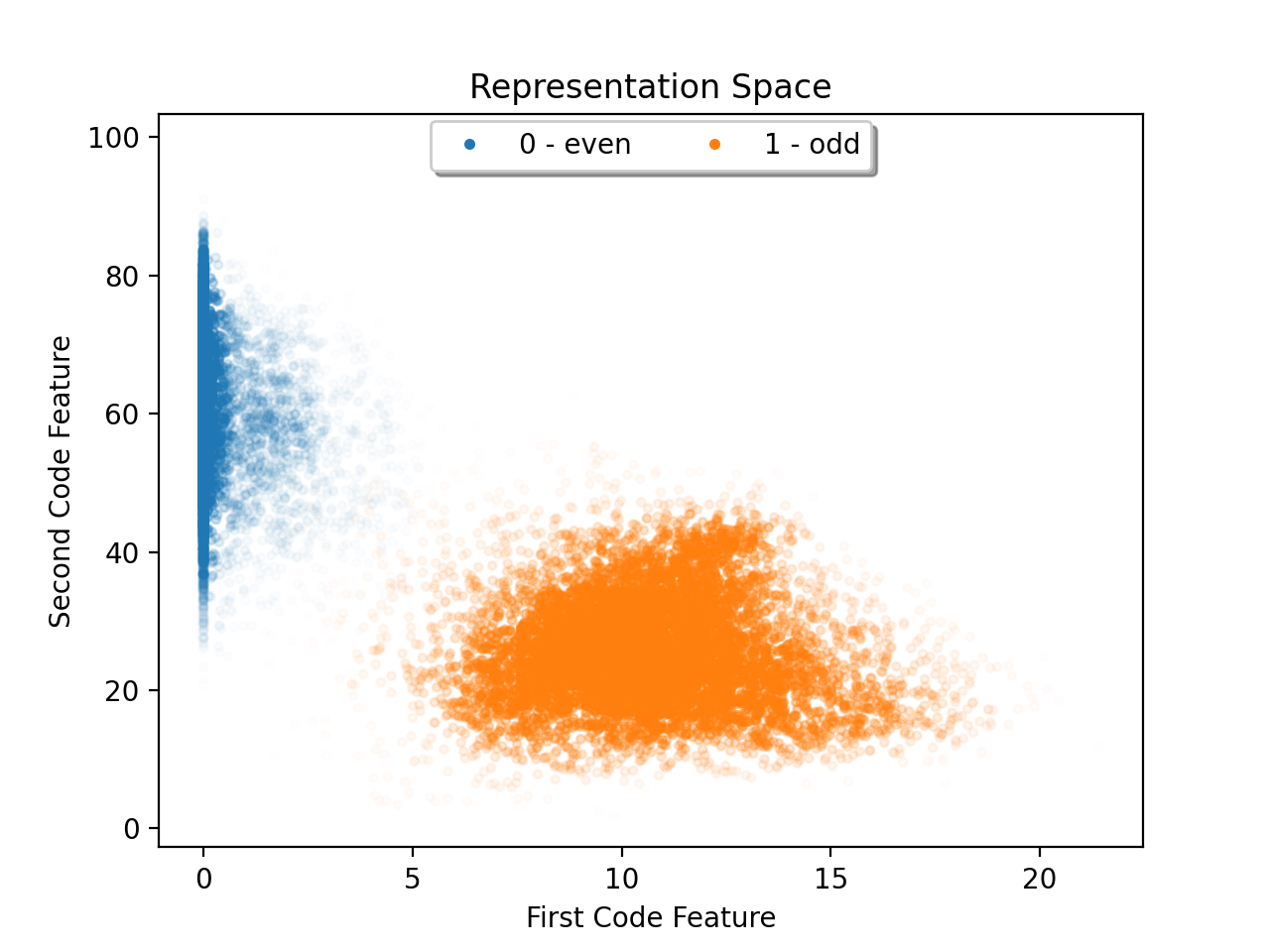}
		\label{fig:sigmoid_embedding}
	} 
	\subfloat[Softmax]
	{
		\includegraphics[width=0.3\linewidth]{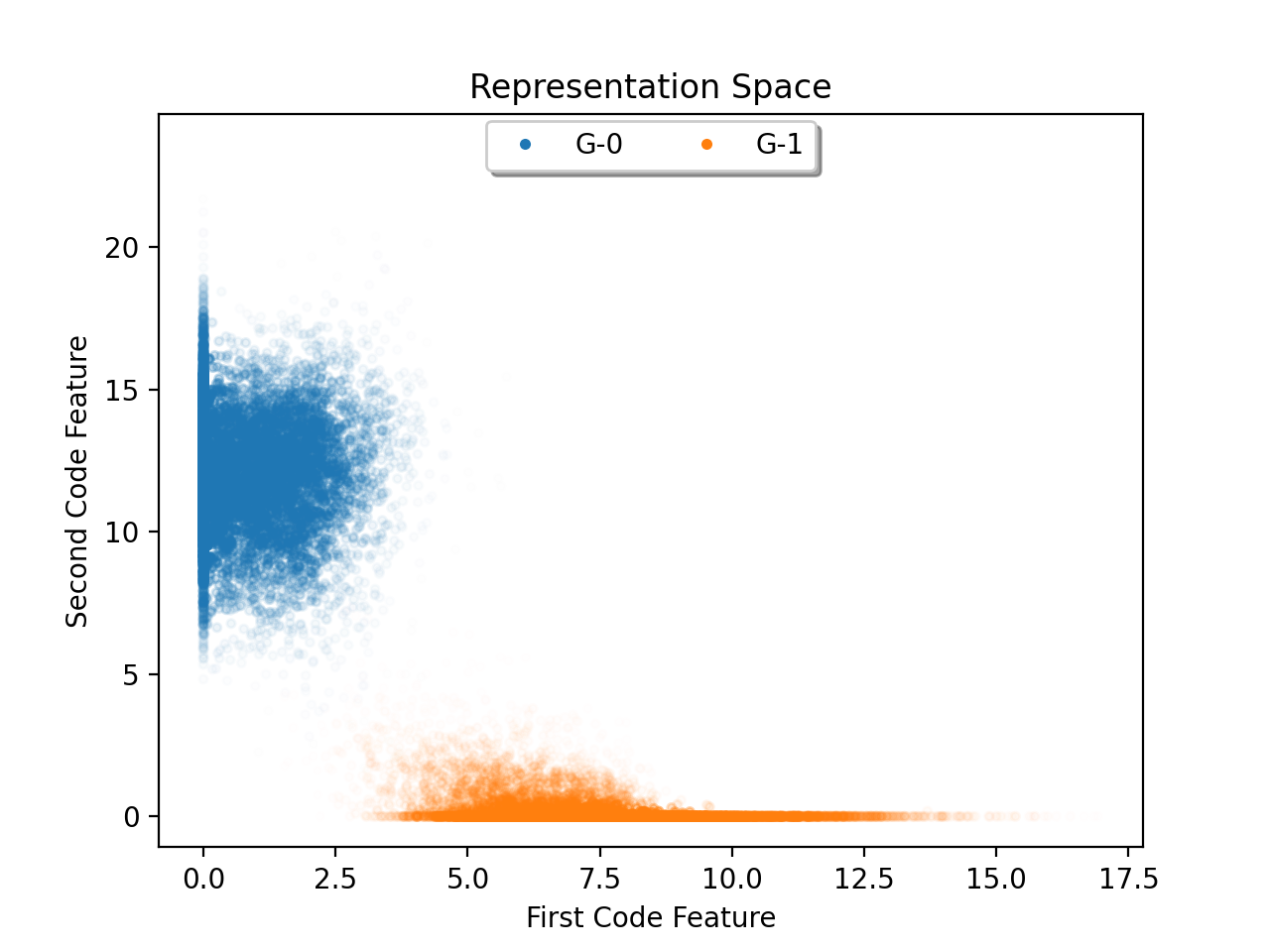}
		\label{fig:softmax_2c_embedding}
	} 
	\subfloat[\dome]
	{
		\includegraphics[width=0.3\linewidth]{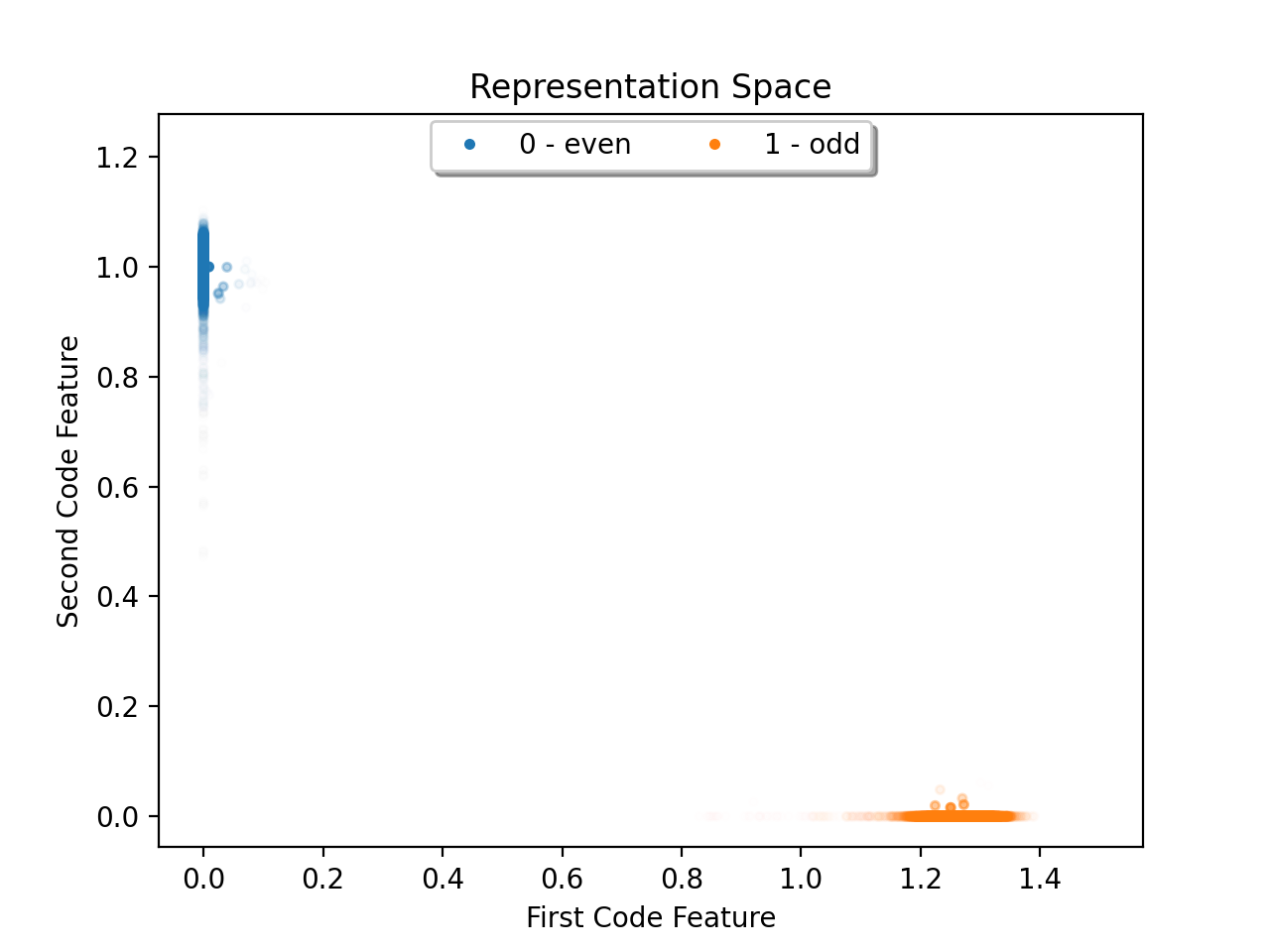}
		\label{fig:dome_embedding}
	} 
	\caption{Embedding spaces resulting from training a binary classification model to classify odd vs even digits in MNIST using different output activation functions. In all cases, the model achitecture is the LeNet \citep{lecun1998LeNet} architecture. The transparency of each point is inversely proportional to its normalized class likelihood estimated using KDE.}
	\label{fig:embeddings_2c}
\end{figure*}

\begin{figure*}[ht]
	\centering
	\subfloat[Sigmoid]
	{
		\includegraphics[width=0.3\linewidth]{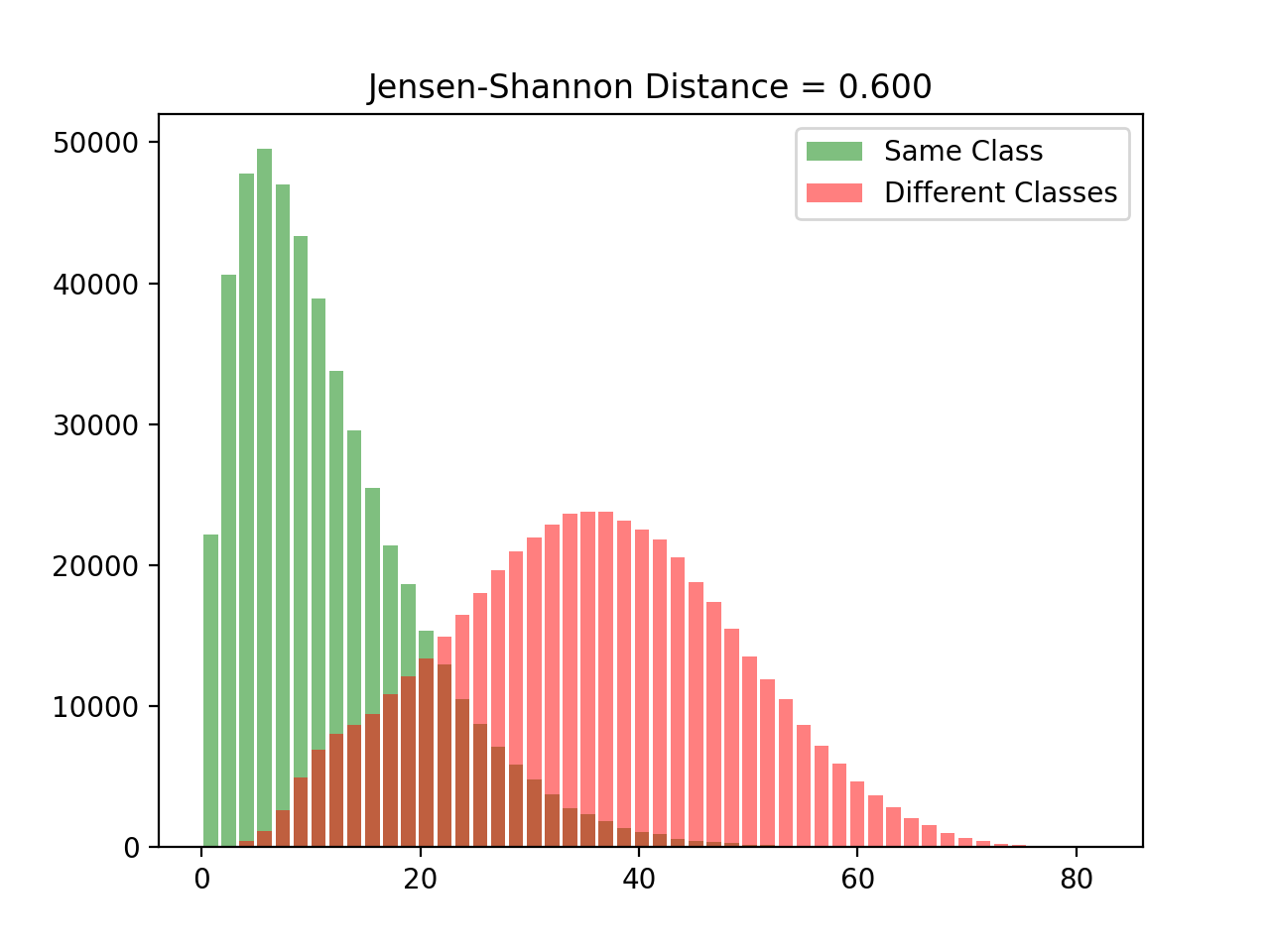}
		\label{fig:sigmoid_jsd}
	} 
	\subfloat[Softmax]
	{
		\includegraphics[width=0.3\linewidth]{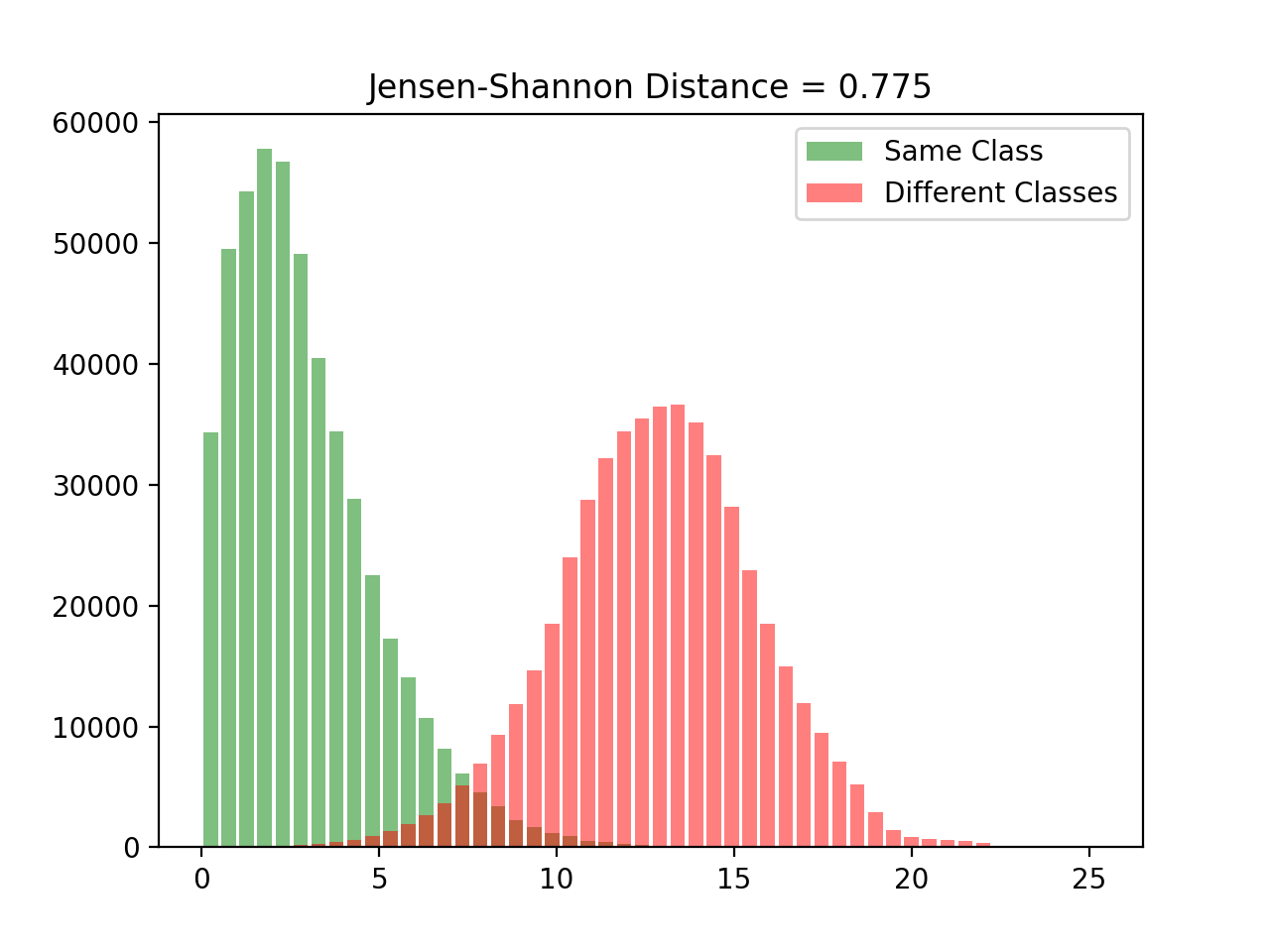}
		\label{fig:softmax_2c_jsd}
	} 
	\subfloat[\dome]
	{
		\includegraphics[width=0.3\linewidth]{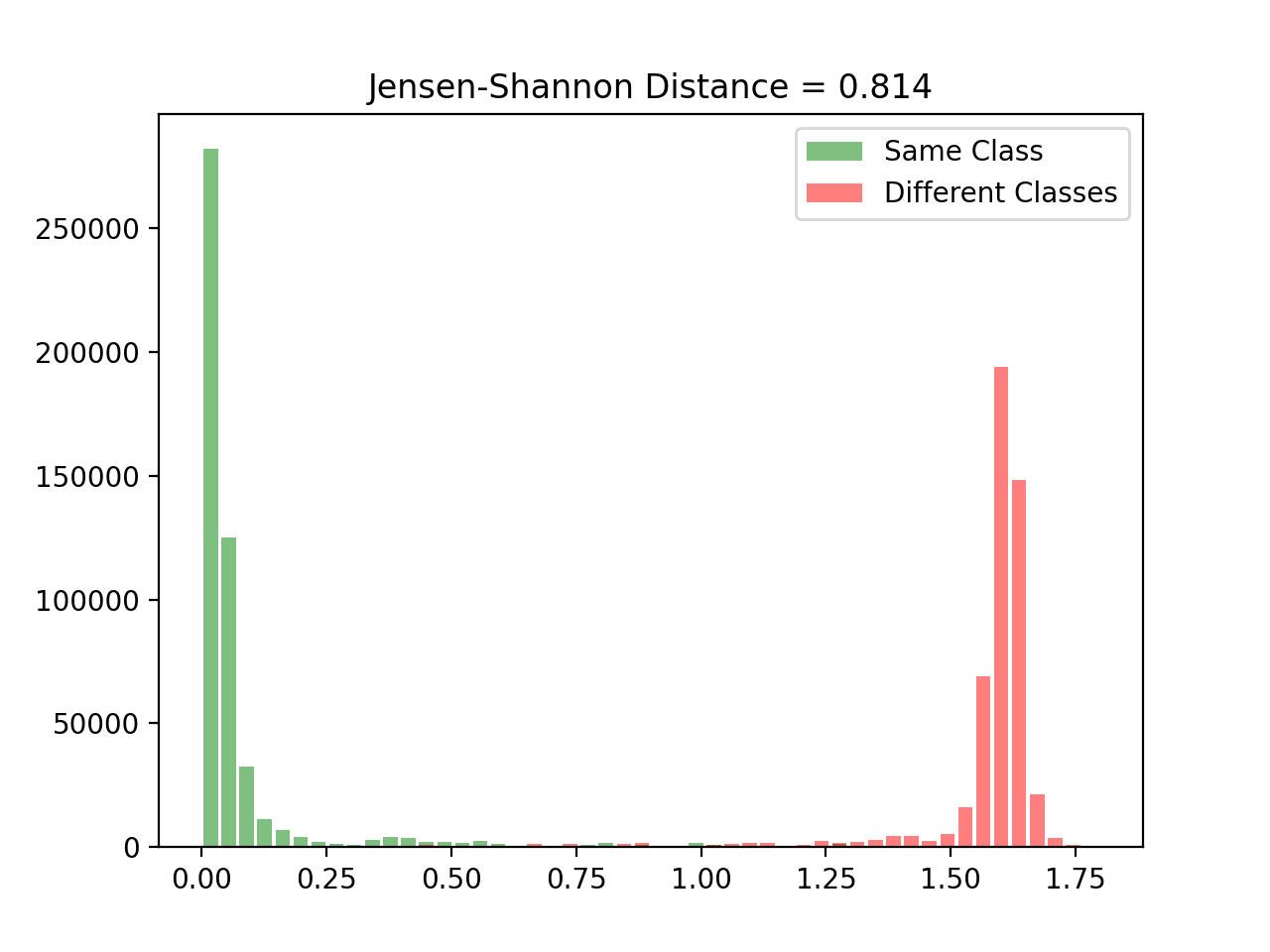}
		\label{fig:dome_jsd}
	} 
	\caption{Inter-class vs intra-class distance distributions over the testing data for the same experiment as in \cref{fig:embeddings_2c} with different output activation functions.}
	\label{fig:jsd_2c}
\end{figure*}

The common characteristic among the three activation functions is that their outputs are almost saturated, at either side of their limits, for infinitely wide ranges of their domains. For example, the tanh function is almost saturated at its maximum value of $1$ in the interval $(3, \infty)$. Since the objective of a binary classification model's training is to make the model's output reach one extreme value for one class and the other extreme value for the other class, optimizing a model for this objective does not impede the logits from growing indefinitely. Consequently, the indefinite growth of the logits has the effect of indefinitely expanding the embedding space.


To examine the expansion of the embedding space, suppose that the logits are the outputs of a linear layer, and, for simplicity, suppose that the linear layer has no bias. In mathematical notation, the $i^{th}$ logit $z_i$ (in binary classification, $i\in \{0, 1\}$) can be expressed as
\begin{equation}
    z_i(\bar{x})=\vec{x}{\,'}\vec{w_i}= \norm{\vec{x}\,} \norm{\vec{w_i}} \cos{\theta} \enspace ,
\end{equation}
where $\vec{x}{\,'}$ is the transpose of the vector from the origin to the input embedding $\bar{x}$, $\vec{w_i}$ is a vector representing the $i^{th}$ column of the linear layer's weight matrix, and $\theta$ is the angle between the two vectors $\vec{x}$ and $\vec{w_i}$. In this case, which is illustrated with a simple 2D embedding space in \cref{fig:2d_logit}, the logit value represents the scaled distance between $\bar{x}$ and a hyperplane ($\phi$ in \cref{fig:2d_logit}) represented by its normal vector $\vec{w_i}$. The scaling factor of such a distance is $\norm{\vec{w_i}}$. Therefore, the logits can grow in absolute values in one of three ways: (1) increasing $\norm{\vec{x}\,}$, which is equivalent to increasing the distance between $\bar{x}$ and the hyperplane ($d$ in \cref{fig:2d_logit}), (2) increasing $\norm{\vec{w_i}}$, which is equivalent to increasing the scaling factor of the distance, or (3) increasing $\cos{\theta}$, which is equivalent to making $\vec{x}$ and $\vec{w_i}$ more parallel or anti-parallel. At a fixed value for the weight vector $\vec{w_i}$, there are only two degrees of freedom to increase $|z_i|$. If the output activation function is any of the ones shown in \cref{fig:tanh_func,fig:sigmoid_func,fig:tanh_func}, once the $|z_i|$ grows beyond a certain value, there will be no noticeable change in the value of the output activation function, and hence there will be no noticeable change in the value of the loss function. In other words, there are hugely different combinations of the values of $\Vec{x}$ and $\theta$ that result in a negligible change in the loss value. As a result, unless proper regularization is deployed, it is typical to have the cloud of points of each class in the embedding space span a large portion of the space. This effect is illustrated in \cref{fig:sigmoid_embedding,fig:softmax_2c_embedding} for the sigmoid and softmax activation functions for the binary classification problem of classifying even versus odd digits in the MNIST dataset. In addition to the unneeded expansion of the embedding space, we can also observe that the inter-class distances can be smaller than the intra-class distances. This phenomenon is depicted in the overlap between the inter-class and intra-class distance distributions in \cref{fig:sigmoid_jsd,fig:softmax_2c_jsd}. We postulate that such lack of compactness in the embedding space makes it easier for an attacker to fabricate an adversarial example due to the small margin separating the two classes.


\section{Difference of mirrored exponential terms}
\label{sec:dome}
To alleviate the problems of output activation functions as outlined in \cref{sec:act_func}, we propose a novel activation function. One objective of our activation function is to have saturated output values only in narrow and finitely bounded ranges of its domain. In this way, the logit values of each class, and hence the corresponding points in the embedding space, will be compactly clustered, and hence the embedding space does not expand indefinitely. We achieve this objective with the activation function in \cref{eq:dome_eq}. We name our function \dome for Difference Of Mirrored Exponential terms because the shape of the function originates from the subtraction of the two exponential terms, which are identical except for the sign in the exponent, i.e. each of them is a vertical mirroring of the other. \Cref{fig:dome_func} shows the shape of \dome when $\mu=\sigma=1$. \Cref{fig:dome_embedding} illustrates the embedding space induced by the function for the same binary classification task used to generate the plots in \cref{fig:embeddings_2c} for the sigmoid and softmax functions. From the figure, it is clear that the embedding space induced by \dome is much more bounded, class-compact, and the intra-class distances are much smaller than the inter-class distances, which is depicted clearly in the inter-class vs intra-class distributions in \cref{fig:dome_jsd}. Indeed, \dome achieves the desired properties it is designed for.

\begin{equation}
    DOME(x) = 0.5\left(1 + e^{-\left(\frac{x-\mu}{\sigma}\right)^2} - e^{-\left(\frac{x+\mu}{\sigma}\right)^2}\right)
    \label{eq:dome_eq}
\end{equation}

\begin{figure}[ht]
	\centering
	\includegraphics[width=1\linewidth]{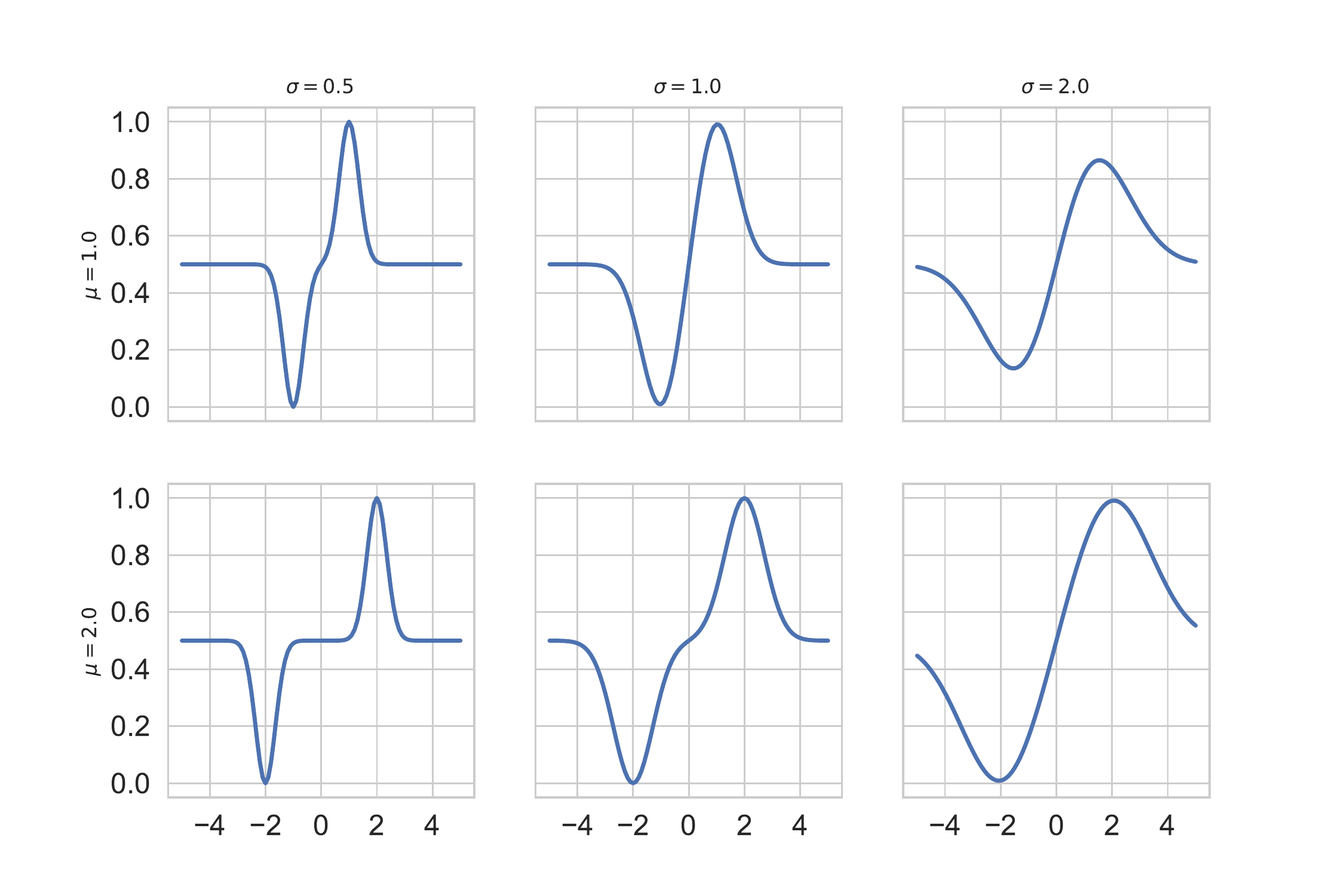}
	\caption{How the shape of the \dome function changes when the relationship between $\mu$ and $\sigma$ does.}
	\label{fig:dome_plots}
\end{figure}

Let's now have a closer look at the behavior of the function. First, because the two exponential terms are bounded within the interval $(0, 1]$, the difference between them is bounded by the interval $(-1, 1)$. After shifting the difference between the two exponential terms by $1$ and scaling by $0.5$, the function becomes bounded within $(0,1)$. Note that the function's output cannot reach its limits due to the subtraction between the two exponential terms. The relationship between the two parameters $\mu$ and $\sigma$ controls how close or faraway the function's output from its limits. The $\mu$ parameter represents the location of the modes of the two mirrored exponential terms. The $\sigma$ parameter represents the breadth of each of these terms. \Cref{fig:dome_plots} illustrates how the shape of the \dome function changes as the relationship between its two parameters, $\mu$ and $\sigma$, changes. The smaller the $\sigma$ value with respect to $\mu$, the relatively narrower the saturation range of the function and the closer the extreme values of its output to the saturation limits. The opposite happens the larger $\sigma$ is with respect to $\mu$. Therefore, if we choose to optimize the two parameters of the function along with other model parameters using binary loss function, such as the binary cross entropy loss, we can expect $\sigma$ to be as small as possible with respect to $\mu$.

\section{Penalized \dome}
\label{sec:pdome}
\begin{figure}[ht]
	\centering
	\includegraphics[width=1\linewidth]{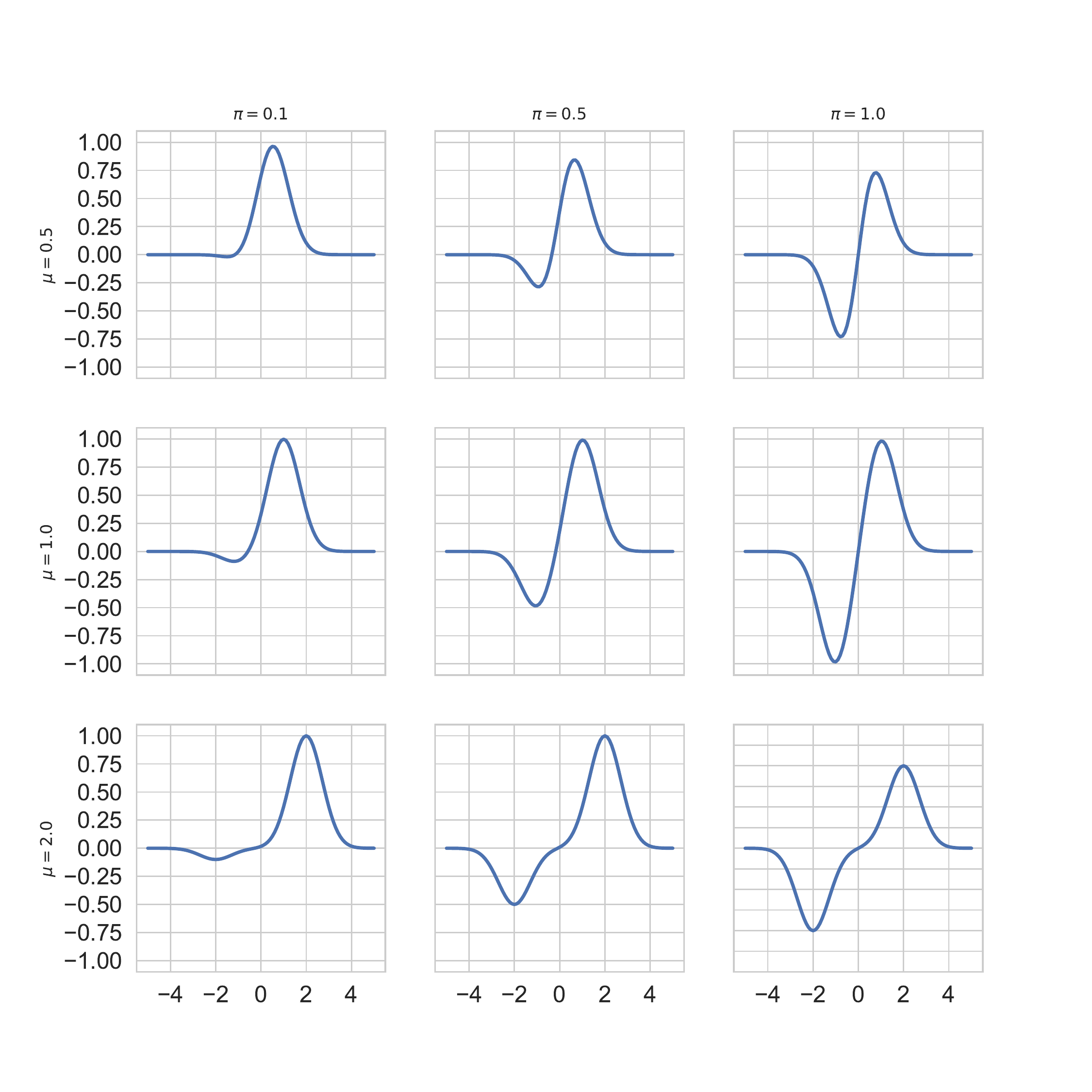}
	\caption{The shapes of the \pdome function with different values of $\mu$ and $\pi$.}
	\label{fig:pdome_plots}
\end{figure}

The \dome activation function can be a proper replacement for sigmoid as an output activation function for binary classification problems. However, the lack of negative output makes it unsuitable for intermediate layers' activation functions where the mix between positive and negative output values is important. Also, it is typical for the intermediate layer activation functions that there is an imbalance between the positive and negative sides. Therefore, we introduce a generalization of \dome, that we call Penalized \dome (\pdome), which offers both properties of having positive and negative outputs, and also having a learnable imbalance between the negative and positive sides. \Cref{eq:dome_n_pos} shows the form of the \pdome function, in which $\pi$ is learnable penalty multiplier for the negative side of the function. This makes $PDOME(x)\in (-pi, 1) \forall x \in R$. When $\pi=1$, \pdome becomes bounded within $(-1,1)$, in which case it becomes similar to the hyperbolic tangent activation function. \Cref{fig:pdome_plots} shows how the shapes of \pdome changes with changing $\mu$ and $\pi$ when $\sigma=1$.

\begin{equation}
    PDOME(x) = e^{-\left(\frac{x-\mu}{\sigma}\right)^2} - \pi e^{-\left(\frac{x+\mu}{\sigma}\right)^2}
    \label{eq:pdome_eq}
\end{equation}

\begin{figure*}[ht]
	\centering
	\subfloat[Embedding space]
	{
		\includegraphics[width=0.3\linewidth]{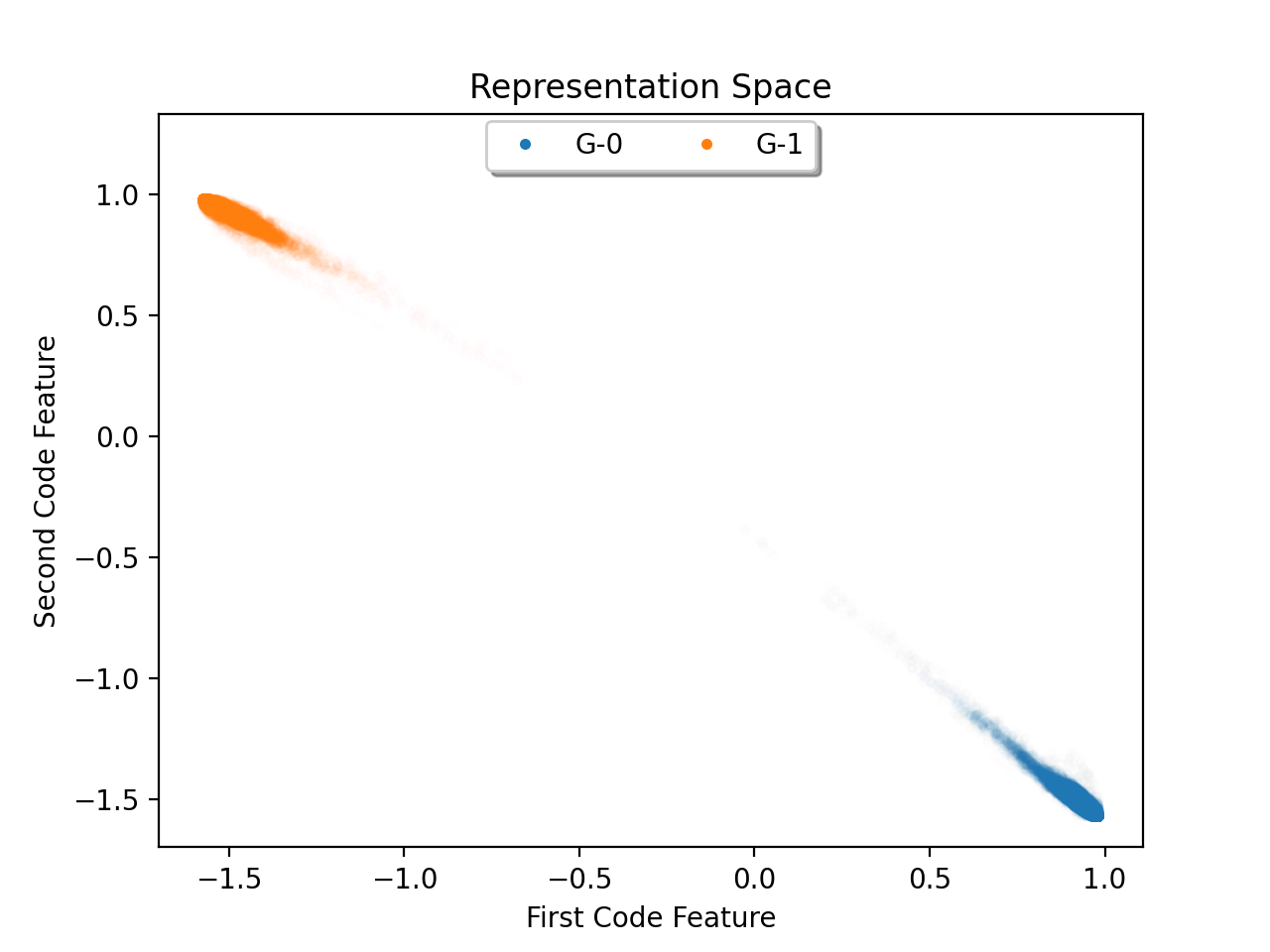}
		\label{fig:pdome_embedding}
	} 
	\subfloat[Learned final \pdome]
	{
		\includegraphics[width=0.3\linewidth]{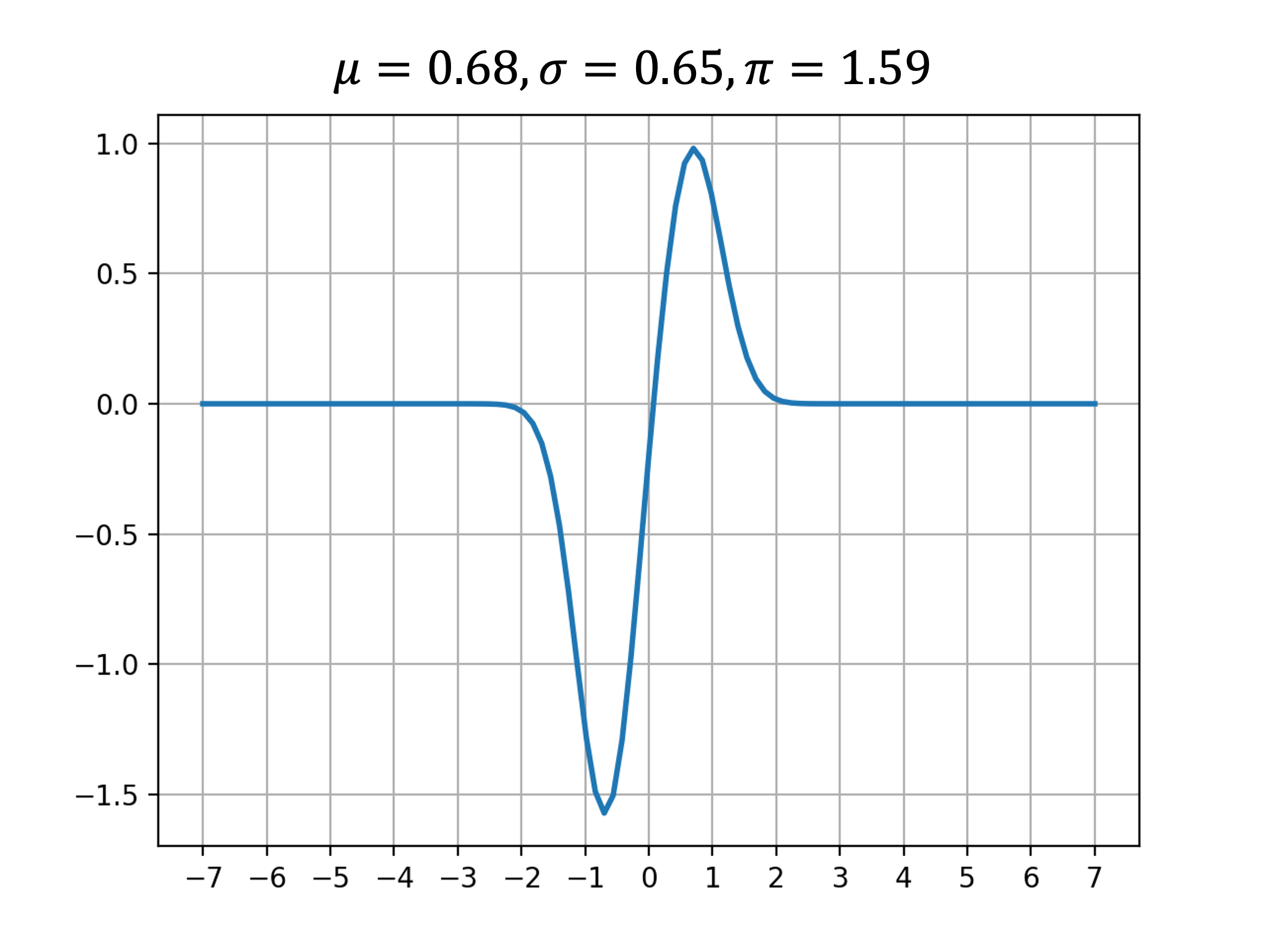}
		\label{fig:pdome_shape}
	} 
	\subfloat[Distance distributions]
	{
		\includegraphics[width=0.3\linewidth]{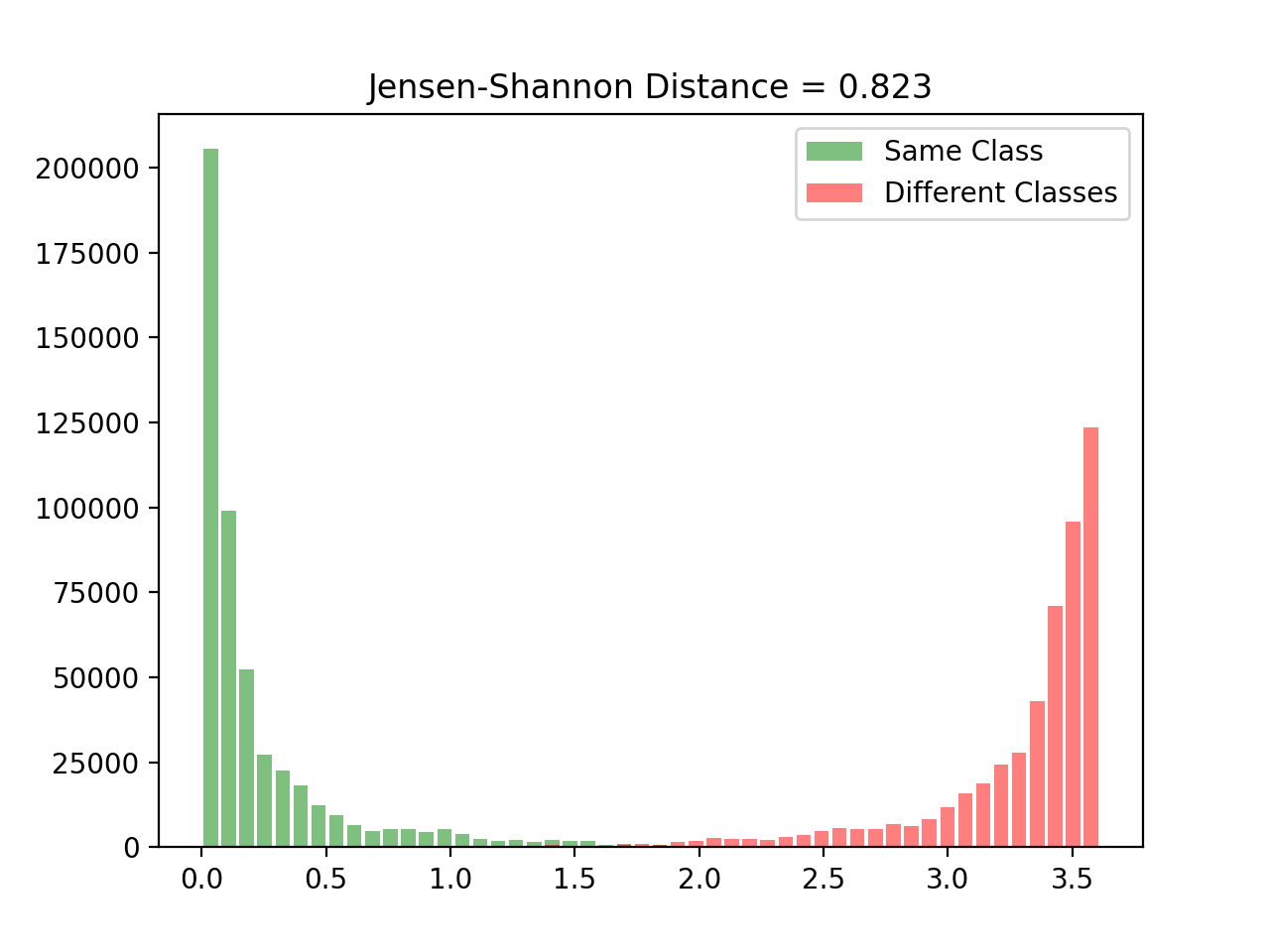}
		\label{fig:pdome_2c_jsd}
	} 
	\caption{The embedding space visualization, the shape of the final-layer's \pdome after training, and the distance distributions when ReLU is replaced with \pdome in the same experiment as in \cref{fig:embeddings_2c}.}
	\label{fig:pdome_trained}
\end{figure*}

We hypothesize that the narrow range of saturated activation values in \pdome results in regularizing the weights of the network without adding explicit regularization or normalization layers, e.g. batch normalization. \Cref{fig:pdome_trained} illustrates the embedding space, the shape of the learned \pdome function, and the distance distributions when the experiment in \cref{fig:embeddings_2c,fig:jsd_2c} is repeated with the ReLU replaced with \pdome. The embedding space appears to be very compact with most of the points concentrated at the peak values of the learned \pdome of the final layer (\cref{fig:pdome_shape}), which is the logits layer. It is also apparent that the points corresponding to each class are closer to one another compared to points belonging to different classes (\cref{fig:pdome_2c_jsd}).

\section{Multi-class generalization of \dome}
\label{sec:mdome}

\begin{figure}[ht]
	\centering
	\includegraphics[width=0.5\linewidth]{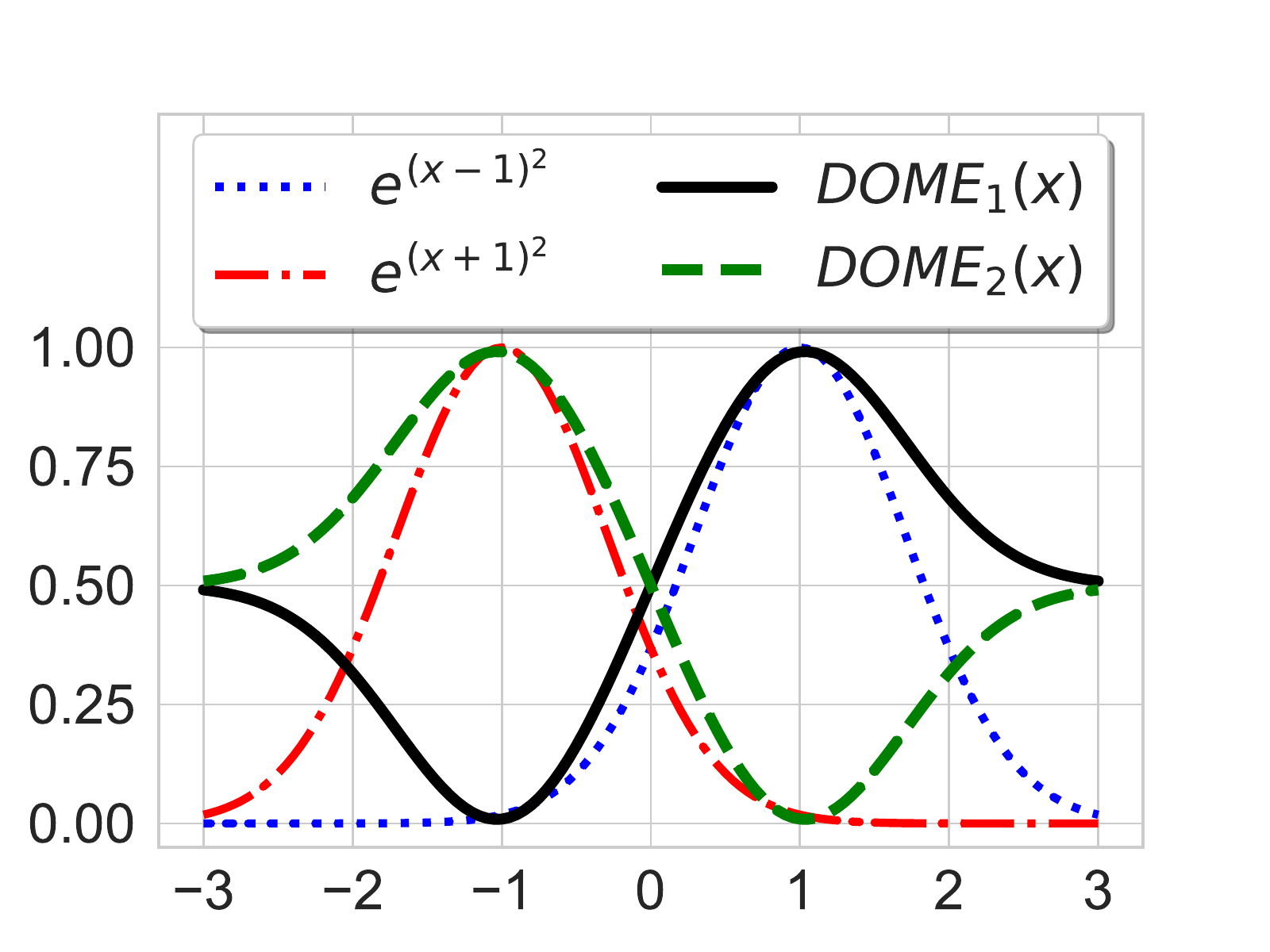}
	\caption{Illustration of the $DOME_1$ and $DOME_2$ (\cref{sec:mdome}) functions and their component exponential terms when $\mu=\sigma=1$.}
	\label{fig:dome_2_anal}
\end{figure}

We showed in \cref{sec:dome} how the \dome function achieves our objectives of class-compactness and regularization to the embedding space in the case of binary classification. We now turn to generalizing \dome's formulation to the multi-class case. First, let's consider a slight variation to the \dome function for the binary classification case, in which instead of producing one output, the function produces two outputs, one for each class, as shown in \cref{eq:dome_2_1} and \cref{eq:dome_2_2},
\begin{align}
    DOME_1(x) = 0.5\left(1 + e^{-\left(\frac{x-\mu e_1}{\sigma}\right)^2} - e^{-\left(\frac{x - \mu e_2}{\sigma}\right)^2}\right),
    \label{eq:dome_2_1}
    \\
    DOME_2(x) = 0.5\left(1 + e^{-\left(\frac{x-\mu e_2}{\sigma}\right)^2} - e^{-\left(\frac{x-\mu e_1}{\sigma}\right)^2}\right),
    \label{eq:dome_2_2}
\end{align}
where $e_1=1$, $e_2=-1$, and is illustrated in \cref{fig:dome_2_anal}. This formulation can be interpreted as that each of the two classes has a point of reference $\mu e_i$ for $i\in{1,2}$. The two points of reference are at the same distance $\mu$ from the origin. The activation value represents the closeness of the input logit $x$ to the reference point of one class compared to the other. Note that the following properties hold
\begin{align}
DOME_1(x) \in (0, 1), &\enspace DOME_2(x) \in (0, 1), \enspace \mbox{and}
\label{eq:dome_2_limits}
\\
DOME_1(x)&+DOME_2(x)=1.
\label{eq:dome_2_sum1}
\end{align}

To extend the \dome's formulation to the multi-class case, we need to draw parallels to the concepts highlighted above. First, we need to have a reference point for each class. Second, we need each reference point to be at the same distance from all others. This last requirement is not applicable in the binary classification case because we only have two reference points. However, one important feature of \dome is that the distance between the two centers is controlled by one parameter, which is $\mu$, such that increasing $\mu$, when the other parameter $\sigma$ is fixed, pushes the reference points away and allows the function's extreme values to become closer to their limits. Another advantage of this requirement is that it makes the formulation of the function symmetric, and hence facilitates its analysis. When we have more than two reference points, we want the same property to hold for all of them at the same time. To achieve this, we need the reference points to occupy the vertices of a regular $n$-simplex for an $n$-class problem. However, an $n$-simplex can only exist in at least an $n-1$ dimensional space. That means, the reference points have to be in an $n-1$ dimensional space or higher. Not only the reference points have to be multi-dimensional, the input to the function has to be mapped to the same multi-dimensional space for the computations to be valid. We assume here that the input to the function has the same multi-dimensional space as the reference points. If this is not the case, the model can be augmented with a projection function, such as a fully connected layer, to perform the mapping from the embedding space to the space of the reference points. All of the above is put together in \cref{eq:dome_n} and \cref{eq:dome_n_exponent}. We refer to this version of \dome as \mdome, for multi-class \dome.
\begin{align}
    DOME^n_i (\bar{x}) & =\frac{n-1}{n}\left(e^{\kappa_i(\bar{x})} -\frac{1}{n-1} \sum\limits_{j\ne i} e^{\kappa_j(\bar{x})} + \frac{1}{n-1} \right)
    \label{eq:dome_n}
    \\
    \kappa_l(\bar{x}) & = -\norm{\frac{\bar{x}-\mu \bar{e_l}}{\sigma}}^2
    \label{eq:dome_n_exponent}
\end{align}
It is easy to see that the 2-class case shown in \cref{eq:dome_2_1} and \cref{eq:dome_2_2} is a special case of the $n$-class case in \cref{eq:dome_n} and \cref{eq:dome_n_exponent} when $n=2$. Given the symmetry of the function, it is also straight forward to prove the following properties of the $n$-class \mdome, which are generalizations of the properties of the $2$-class case in \cref{eq:dome_2_limits} and \cref{eq:dome_2_sum1}.
\begin{align}
DOME^n_i(\bar{x}) \in \left(\frac{2-n}{n}, 1\right) 
\label{eq:dome_n_limits}
\\
\sum\limits_{i\in[n]} DOME^n_i(\bar{x}) =1
\label{eq:dome_n_sum1}
\end{align}

In contrast to the binary case, the lower bound of the \mdome function can be negative. The lower bound in \cref{eq:dome_n_limits} approach $-1$ as $n$ approaches infinity. However, this lower bound is loose and in practice the function never approaches it. In practice, we normalize the function's values to be all non-negative as shown in \cref{eq:dome_n_norm} and \cref{eq:dome_n_pos}.

\begin{align}
\hat{DOME}^n_i(\bar{x}) = \frac{\Tilde{DOME}^n_i(\bar{x})}{\sum\limits_{i\in[n]} \Tilde{DOME}^n_i(\bar{x})} \enspace, \text{ where}
\label{eq:dome_n_norm}
\\
\Tilde{DOME}^n_i(\bar{x}) = {DOME}^n_i(\bar{x}) - \min\limits_{k\in [n]}\left( \min\left({DOME}^n_k(\bar{x}), 0\right) \right)
\label{eq:dome_n_pos}
\end{align}

\begin{figure}[ht]
	\centering
	\subfloat[\mdome's embedding space]
	{
		\includegraphics[width=0.47\linewidth]{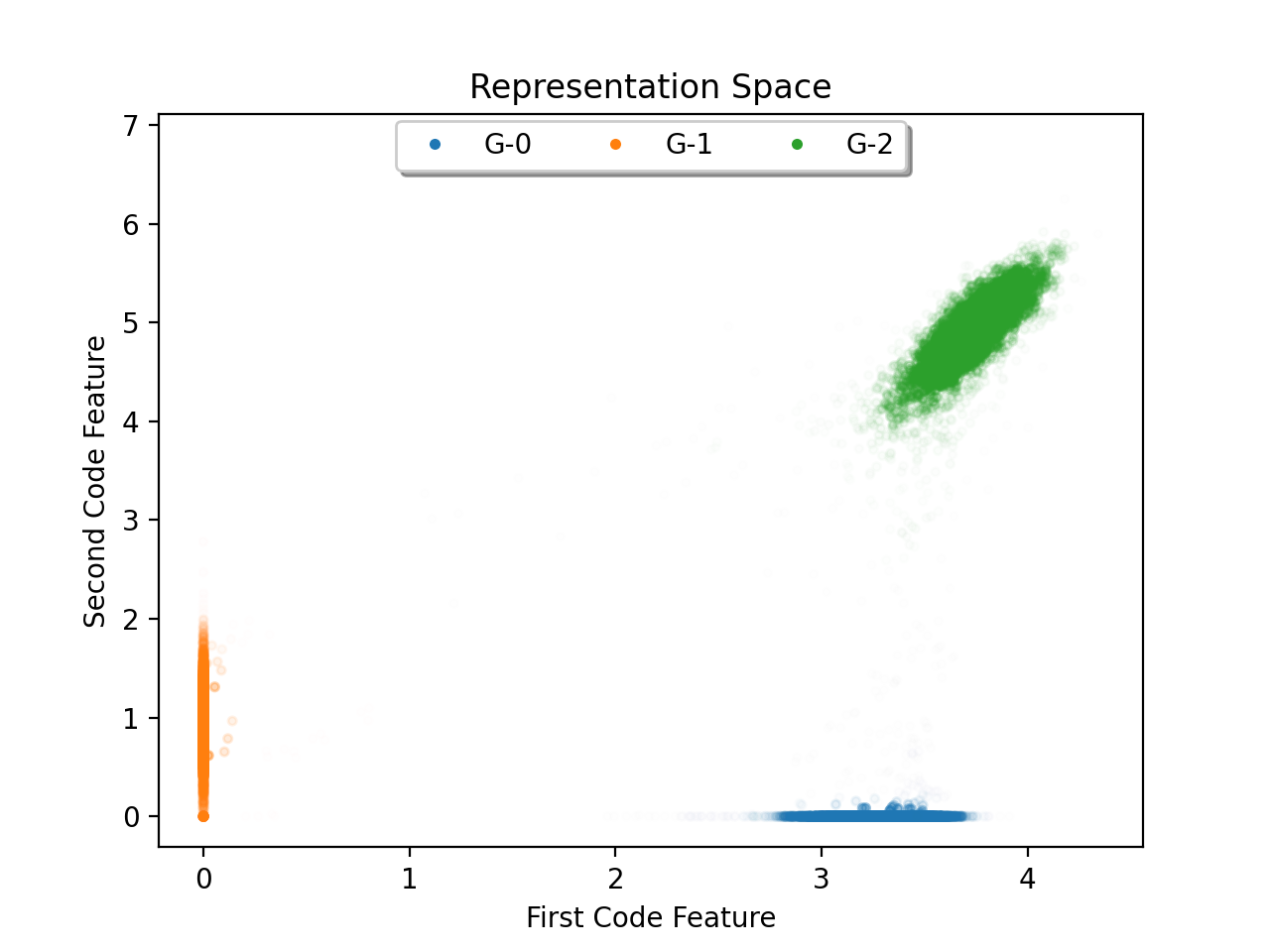}
		\label{fig:mdome_embedding}
	} 
	\subfloat[Softmax' embedding space]
	{
		\includegraphics[width=0.47\linewidth]{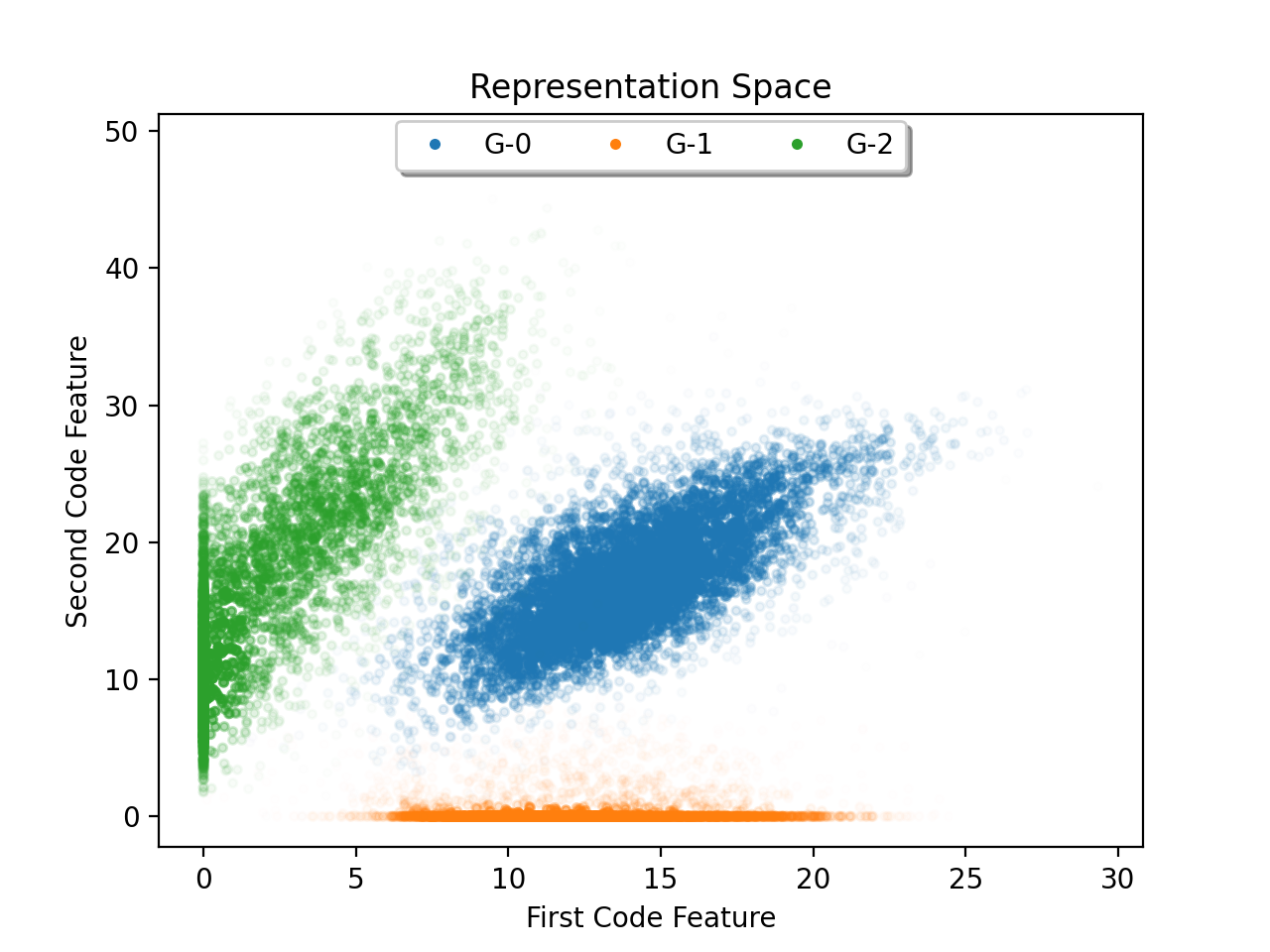}
		\label{fig:softmax_3c_embedding}
	} 
	\\
	\subfloat[\mdome's distance distributions]
	{
		\includegraphics[width=0.47\linewidth]{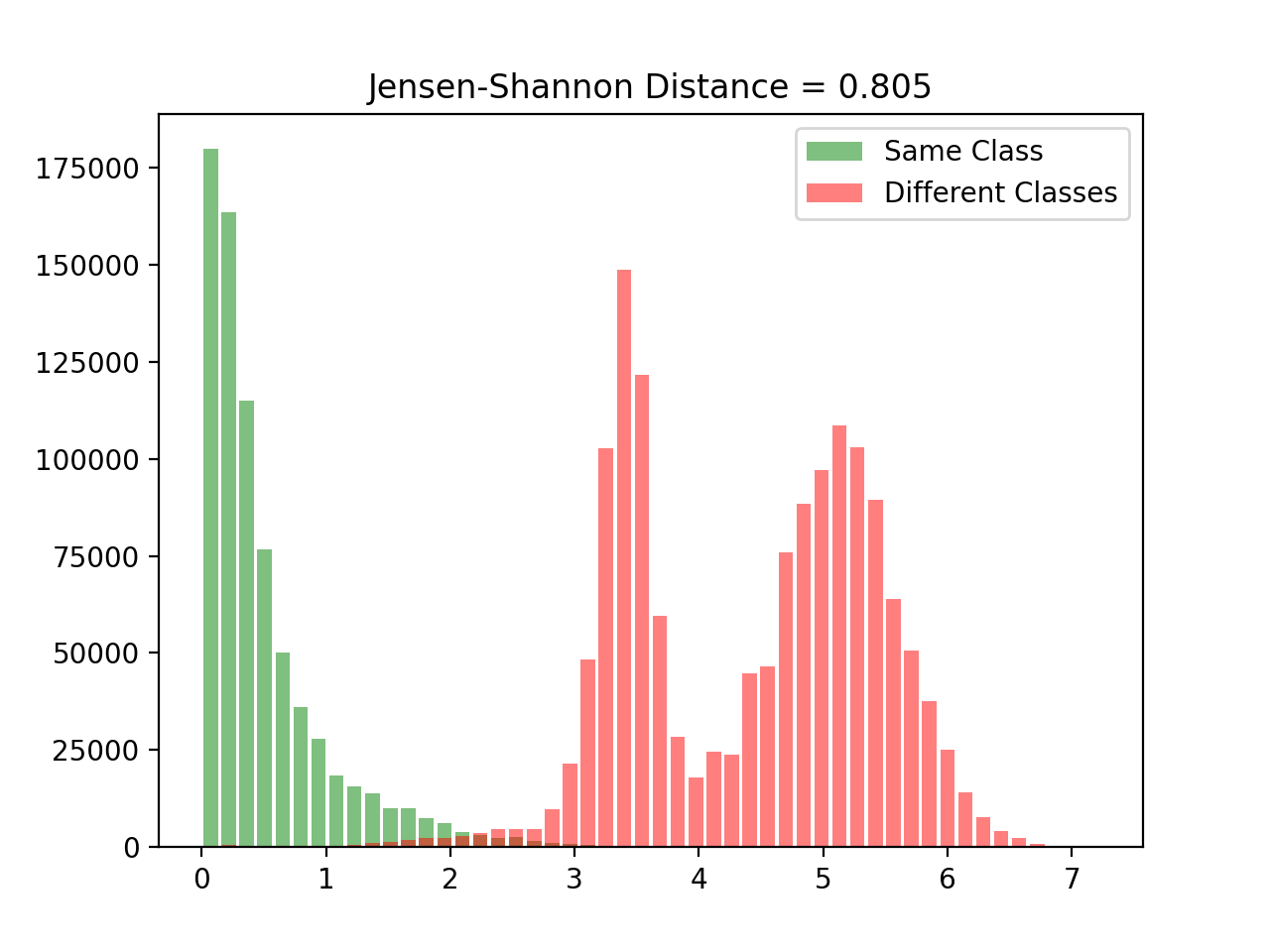}
		\label{fig:mdome_jsd}
	} 
	\subfloat[Softmax' distance distributions]
	{
		\includegraphics[width=0.47\linewidth]{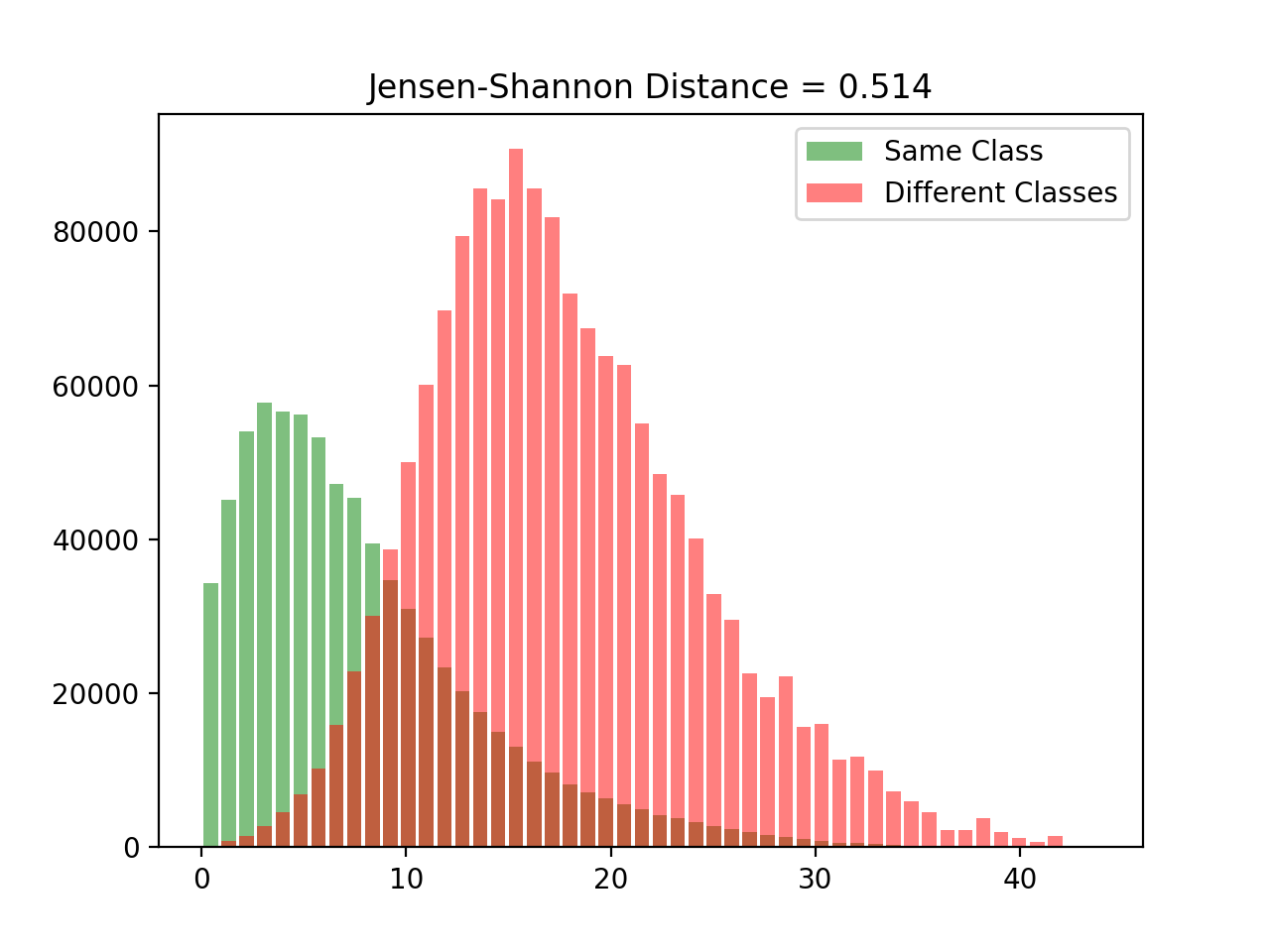}
		\label{fig:softmax_3c_jsd}
	} 
	\caption{Embedding spaces and inter-class vs intra-class distance distributions over the testing data for an experiment similar to the one in \cref{fig:embeddings_2c}, but with three classes, with different output activation functions.}
	\label{fig:embeddings_jsd_3c}
\end{figure}

We repeated the same experiments used in \cref{fig:embeddings_2c,fig:jsd_2c,fig:pdome_trained} on the MNIST data using the LeNet architecture \citep{lecun1998LeNet} with a 2D embedding space, but with 3 classes instead of 2, where the class is the $digit~\text{mod}~3$. The resulting embedding spaces and inter-class/intra-class distance distributions for the softmax and \mdome activation functions are shown in \cref{fig:embeddings_jsd_3c}. The plots underscores the significant difference between the \mdome and softmax in terms of class compactness and regularization of the embedding space.

\section{Experimental evaluation}
\label{sec:exps}
We evaluate the \pdome and \mdome functions on three datasets, which are MNIST \citep{lecun1998LeNet}, Fashion-MNIST \citep{xiao2017FashionMNIST}, and CIFAR-10 \citep{krizhevskyLearningMultipleLayers2009}. For each dataset, we trained a ResNet-18 network using the fast adversarial training (FAT) framework \citep{wongFastBetterFree2019}. We used the default optimizer (SGD) with momentum of $0.9$ and learning rate scheduler (linear cyclic learning rate scheduler), and default parameters $\epsilon=\frac{8}{255}$ and $\alpha=\frac{10}{255}$. We set the number of epochs to 30. We used grid search to tune the learning rate, weight decay, and whether Nesterov momentum is used or not, for each model and dataset. We empirically found that the mean-square-error loss works best with \mdome. Therefore, it is used in all our \mdome experiments. The initial $\mu$ for the \dome functions was always set to $1$. For \pdome, the initial $\sigma$ and $\pi$ were set to $1$ and $0.1$, respectively. For \mdome, the initial $\sigma$ was set to $5$. \Cref{tab:hyper} shows the final hyper parameters used to produce the results in this paper.

\begin{table}[ht]
    \footnotesize
    \centering
    \begin{tabular}{lccc}
    \toprule
         &  Baseline & \pdome & \mdome \\
    \midrule
    MNIST        & ${0.1,0.0005,F}$ & ${0.02,0,F}$ & ${0.5,0.0005,T}$  \\
    Fashin-MNIST & ${0.1,0.0005,F}$ & ${0.1,50.0005,F}$ & ${0.1,0.0005,T}$  \\
    CIFAR-10     & ${0.2,0.0005,F}$ & ${0.02,0.0025,F}$ & ${0.5,0.0005,T}$  \\
    \bottomrule
    \end{tabular}
    \caption{Hyper parameters for each experiment. Values are ordered as learning rate, weight decay, Nesterov momentum (\textbf{T}rue/\textbf{F}alse).}
    \label{tab:hyper}
\end{table}

\Cref{tab:accuracies} shows the benign and adversarial accuracy values for the three datasets comparing the baseline model, which uses ReLU as the layer-wise non-linear activation function and softmax as the output activation function, and two other models, one replaces the ReLU in the baseline with \pdome, and the other replaces the softmax with \mdome. The adversarial accuracy is based on two white-box attacks. The first is the PGD attack \citep{madryDeepLearningModels2018} with 50 iterations and 10 random restarts. The other is the Auto-Attack \citep{croceReliableEvaluationAdversarial2020b}. For the latter, we use the implementation in the adversarial-robustness-toolbox (ART) \citep{art2018}, which applies a bundle of four attacks: two variants of PGD, DeepFool \citep{Moosavi-Dezfooli_2016_CVPR} and SquareAttack \citep{andriushchenkoSquareAttackQueryEfficient2020}. We used the default setting of ART except for setting the random restart for SquareAttack to 1 instead of 5 to keep the running time in a reasonable range. The epsilon and epsilon step, for all attacks, were set to $\frac{8}{255}$ and $\frac{2}{255}$, respectively.

Experimenting with PGD alone might seem redundant since it is already included in the Auto-Attack's bundle. However, the standalone PGD in our experiments is different because it uses the loss function used to train the model, which is the MSE loss in the case of \mdome. However, the PGD in Auto-Attack's bundle uses other two custom loss functions. These loss functions expect the logits to be the output of the model. Since \mdome outputs probabilities, we experimented with three different variants of \mdome and reported the results based on the union of the found adversarial examples with the three variants. The first variant uses \mdome's probabilities as logits. The second two variants use the two functions in \cref{eq:dome_adaptive1,eq:dome_adaptive2}. Therefore, Auto-Attack in our case is considered an adaptive attack \citep{tramerAdaptiveAttacksAdversarial2020}, which is designed based on the knowledge about the defense.

\begin{align}
{DOME-LOGIT1}^n_i(\bar{x}) & = -\norm{\frac{\bar{x}-\mu \bar{e_i}}{\sigma}}^2
\label{eq:dome_adaptive1}
\\
{DOME-LOGIT2}^n_i(\bar{x}) & = \bar{x} \cdot \mu \bar{e_i}
\label{eq:dome_adaptive2}
\end{align}
where $\cdot$ is the dot product operator.

The results show the consistent advantage of the \mdome compared to the baseline in terms of adversarial accuracy for the three datasets and the two attacks. In all the cases, \mdome supersedes the baseline except for one case in which it comes as a close second. For the Fashion-MNIST dataset, the \pdome emerges as the most robust under the two attacks with \mdome as a close second. The extra robustness comes at the expense of degraded benign accuracy, which is mostly minor except for the case of the CIFAR-10 dataset.

\begin {table*}
\footnotesize
\centering
\caption{Classification accuracy on benign and adversarial examples for models trained using FAT \citep{wongFastBetterFree2019}.}
\label{tab:accuracies}
\begin{tabular}{l|ccc|ccc|ccc}
\toprule
\multirow{2}{*}{Dataset} & \multicolumn{3}{c}{Benign} & \multicolumn{3}{c}{PGD-50x10} & \multicolumn{3}{c}{Auto-Attack}
\\
 & Baseline & \pdome & \mdome & Baseline & \pdome & \mdome & Baseline & \pdome & \mdome 
 \\
\midrule
MNIST           & $\underline{99.70}$ & ${99.63}$ & $\pmb{99.74}$ & $\underline{99.53}$ & ${99.52}$ & $\pmb{99.55}$ & $\pmb{99.53}$ & ${99.51}$ & $\underline{99.52}$ \\
Fashion-MNIST   & $\pmb{92.86}$ & $\underline{92.71}$ & ${92.38}$ & ${86.82}$ & $\pmb{88.26}$ & $\underline{88.08}$ & ${85.12}$ & $\pmb{88.09}$ & $\underline{87.98}$ \\
CIFAR-10        & $\pmb{83.48}$ & ${77.50}$ & $\underline{79.53}$ & $\underline{45.99}$ & ${41.83}$ & $\pmb{46.77}$ & $\underline{43.79}$ & ${42.27}$ & $\pmb{44.09}$ \\
\bottomrule
\end{tabular}
\end {table*}

\section{Conclusion}
\label{sec:conc}
We introduced a novel activation function dubbed \dome, for Difference of Mirrored Exponential terms. \dome has a couple of interesting properties. First, it naturally induces intra-class compactness and inter-class divergence to the embedding space without using any special loss. Second, it has a self-regularization property, which means that, training a network with \dome without any other regularization mechanism, the embedding space does not excessively grow, which is typically the case with traditional output activation functions. \dome can be extended to the multi-class case, which we call \mdome. It can also be generalized to take more arbitrary shapes amenable to non-linear activation for internal layers, which we call \pdome, for Penalized \dome. Our experiments show that the \dome variant enhances the robustness against white-box evasion attacks.

\section*{Acknowledgments}

This research is based upon work supported 
by the Office of the Director of National Intelligence (ODNI), Intelligence Advanced Research Projects Activity (IARPA), under contract number 2017-17020200005, and 
by the Defense Advanced Research Projects Agency (DARPA), under cooperative agreement number HR00112020009. The  views  and  conclusions  contained  herein should not be interpreted as necessarily representing the official policies or endorsements, either expressed or  implied,  of  
the  ODNI,  IARPA, 
DARPA or the  U.S.  Government. The U.S. Government is authorized to reproduce and distribute reprints for governmental purposes notwithstanding any copyright notation thereon.

\ifprl
\bibliographystyle{model2-names.bst}
\else
\bibliographystyle{acm}
\fi
\bibliography{refs}

\end{document}